\documentclass[a4paper,fleqn]{cas-dc}

\usepackage{soul} 
\usepackage{color, xcolor}

\usepackage{natbib}
\usepackage{apalike}

\usepackage{amsthm}

\usepackage{subfigure}
\usepackage{tabularx}
\usepackage{tikz}
\usepackage{pgfplots}
\usepackage[edges]{forest}
\usepackage{adjustbox}

\usepackage{makecell}

\usepackage{tabularx}

\newtheorem{definition}{Definition}[section]
\counterwithout{definition}{section}

\counterwithout{principle}{section}

\usepackage{subfigure}
\usepackage{multirow}
\usepackage{amsmath} 

\usepackage{amssymb} 
\usepackage{rotating}

\usepackage{chngcntr}
\counterwithout{definition}{section}

\usepackage{makecell}
\usepackage{algorithm, algorithmic}

\usepackage[switch]{lineno}
\usepackage{ulem}

\begin{document}

\title[mode = title]{Pure Node Selection for Imbalanced Graph Node Classification}

\shorttitle{Pure Node Selection for Imbalanced Graph Node Classification}
\shortauthors{F. Zeng \textit{et al.}}

\author[1]{Fanlong Zeng}
\ead{flzeng1@stu.jnu.edu.cn}
\address[1]{School of Intelligent Systems Science and Engineering, Jinan University, Zhuhai 519070, China}

\author[1]{Wensheng Gan}
\cortext[cor1]{Corresponding author}
\ead{wsgan@jnu.edu.cn}
\cormark[1]

\author[2]{Jiayang Wu}
\ead{jywu1@stu.jnu.edu.cn}
\address[2]{College of Cyber Security, Jinan University, Guangzhou 510632, China}

\author[3]{Philip S. Yu}
\ead{psyu@uic.edu}
\address[4]{Department of Computer Science, University of Illinois Chicago, Chicago 60607, USA}


\maketitle
\begin{abstract}
    The problem of class imbalance refers to an uneven distribution of quantity among classes in a dataset, where some classes are significantly underrepresented compared to others. Class imbalance is also prevalent in graph-structured data. Graph neural networks (GNNs) are typically based on the assumption of class balance, often overlooking the issue of class imbalance. 
    In our investigation, we identified a problem, which we term the Randomness Anomalous Connectivity Problem (RACP), where certain off-the-shelf models are affected by random seeds, leading to a significant performance degradation. To eliminate the influence of random factors in algorithms, we proposed PNS (\underline{P}ure \underline{N}ode \underline{S}ampling) to address the RACP in the node synthesis stage. 
    Unlike existing approaches that design specialized algorithms to handle either quantity imbalance or topological imbalance, PNS is a novel plug-and-play module that operates directly during node synthesis to mitigate RACP. Moreover, PNS also alleviates performance degradation caused by abnormal distribution of node neighbors.
     We conduct a series of experiments to identify what factors are influenced by random seeds. 
    Experimental results demonstrate the effectiveness and stability of our method, which not only eliminates the effect of unfavorable random seeds but also outperforms the baseline across various benchmark datasets with different GNN backbones. Data and code are available at \href{https://github.com/flzeng1/PNS}{https://github.com/flzeng1/PNS}.
\end{abstract}

\section{Introduction}\label{sec:Introduction}

Node classification is a crucial task in the graph domain \citep{cook2006mining, ju2024comprehensive}. Graphs are primarily used to represent complex relationship networks between entities. In a graph, nodes typically represent individual entities, which can be persons, objects, organizations, events, or other concepts. For instance, a social network can be effectively modeled as a graph, where each node corresponds to a unique user, and the edges between nodes represent the relationships between them.  
There are many real-world demands for node classification \citep{wu2022graph, yang2021consisrec, wu2025graph, wu2025adkgd}, such as identifying malicious users on social networks \citep{ying2018graph}. In protein-protein interaction networks, this technique is utilized to predict the functional annotation or classification of a protein \citep{jumper2021highly}. As a powerful tool for tackling node classification tasks, graph neural networks (GNNs) have consistently demonstrated their effectiveness in various graph-related applications \citep{wang2022powerful, wu2019simplifying}. Currently, GNNs operate under the assumption that classes are balanced, overlooking the potential issue of class imbalance \citep{zhao2021graphsmote, liu2023imbens}. The class imbalance problem arises when the quantity of data for different classes is unevenly distributed, leading to an imbalance in the amount of data available for training. The graph class imbalance problem belongs to the broader class imbalance problem, but has some notable differences. It is characterized not only by a disparity in quantity but also by topological structure \citep{park2021graphens, song2022tam}. The graph data is class-imbalanced in reality \citep{shchur2018pitfalls, galke2023lifelong}. For Netflix, the number of customers is much larger than the number of videos.  Using GNNs on class-imbalanced graphs can lead to underrepresentation of the minority class, negatively impacting performance and hindering generalization \citep{song2022tam, park2021graphens}. Therefore, it is essential to explore solutions to the problem of graph imbalance.

\begin{figure}
    \centering
    \vspace{0.6cm} 
    \includegraphics[scale=0.45]{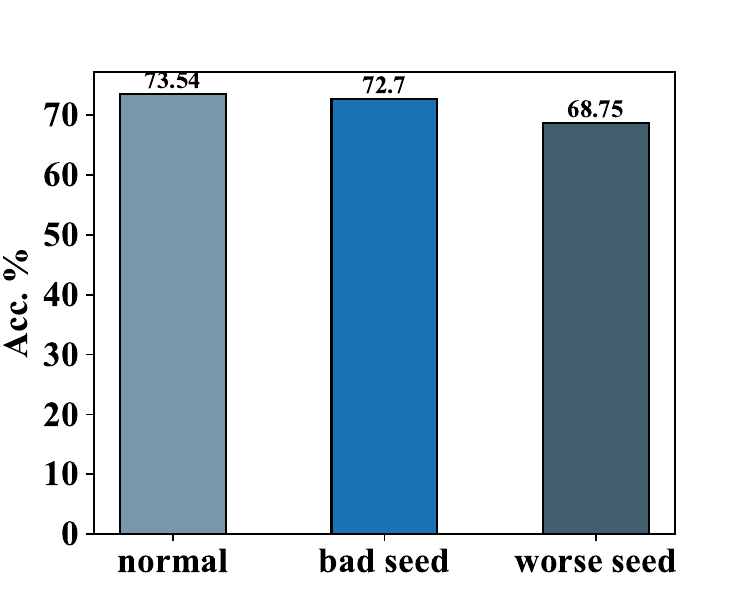}
    \caption{The RACP in GraphENS \citep{park2021graphens}, which can be intuited by examining the accuracy performance of the model under different random seeds. The experiment is GraphENS training in the Cora-LT dataset with GCN as a backbone with distinct random seeds. The ``normal''  denotes the baseline performance where the random seed maintains standard performance without inducing RACP. While ``bad seed'' and ``worse seed'' denote the degradation of the model performance by RACP to different degrees. The normal performance is denoted by the first bar, while the other two bars represent the performance of the model with different random seeds, resulting in decreased performance.}
    \label{fig: rac problem pic}
\end{figure}

Research on graph imbalance is scarce, but a few methods have been proposed. These methods can be categorized into node synthesis methods \citep{chawla2002smote, li2023graphsha, park2021graphens} and loss-modified methods \citep{chen2021topology, liu2021tail, song2022tam}. Node synthesis methods aim to balance data distribution by augmenting the minority class. Loss-modified methods alter the loss function to favor the minority class. In this paper, node synthesis is a key aspect of our approach. We reconsider the graph imbalance problem from the perspective of reducing the sampling boundary of nodes, supported by an empirical study in Fig. \ref{fig: rac problem pic}, which illustrates the performance degradation caused by randomness in GraphENS \citep{park2021graphens}, a graph imbalance method. The random performance degradation problem refers to the decline in GraphENS performance under different random seeds. Our experiments revealed that poor random seeds led to the generation of more abnormal nodes, which in turn adversely affected the performance of GraphENS.

The primary challenge in addressing the random performance degradation problem lies in mitigating the impact of random seeds on the model's performance. Despite the clarity of the goal, determining the root cause of the randomness anomalous connectivity problem remains a formidable challenge. The factors that random seeds can influence are endless. Therefore, to mitigate the impact of randomness on the existing model's performance, we need to conduct experiments to identify the source of randomness and attempt to eliminate it without compromising the model's overall performance.

\begin{figure}
    \centering
    \includegraphics[scale=0.65]{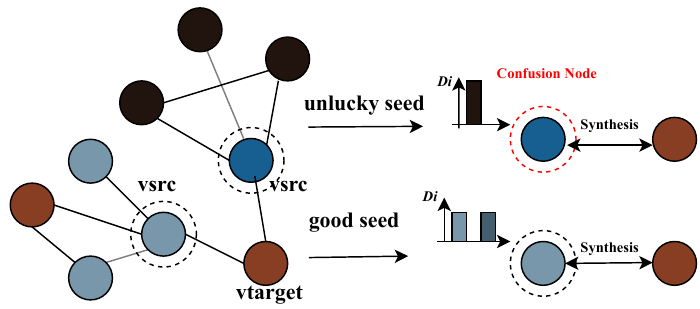}
    \caption{It illustrates how seed-induced sampling variance affects synthesized node quality. The color of the node denotes a different class. During node sampling, the model probabilistically selects node pairs for feature mixing, where random seed initialization influences sampling outcomes. For example, in the upper part of the figure, ``unlucky seed'' would induce the model to sample a source node with anomalous connectivity to synthesize a new node with the target node, where the new node would potentially be a confusion node. With the ``good seed'', the synthesized node is akin to a normal node. By incorporating a confusion node into the model, the results of a mixed node can corrupt the message-passing mechanism of a GNN, ultimately compromising the model's performance \citep{song2022tam}.}
    \label{fig: rac problem}
\end{figure}

To address the challenge, we propose PNS for \underline{P}ure \underline{N}ode \underline{S}ampling, designed to mitigate the impact of randomness and enhance the model's overall performance. In our experiment, the random performance degradation problem arises when the model selects nodes with anomalous connectivity based on the random seed—a phenomenon we term the randomness anomalous connectivity problem (RACP), like in Fig. \ref{fig: rac problem}. To prevent this situation from arising, we propose a new module named pure node sampling (PNS). We adjust the node selection boundaries based on the distribution of one-hop neighbor labels to determine which nodes should be included in the sampling boundaries. Furthermore, PNS can mitigate performance degradation caused by the abnormal distribution of node neighbors. We evaluate our method on many different datasets, including citation networks such as Cora, Citseer, and PubMed \citep{sen2008collective}, as well as Amazon purchase networks like Amazon-Computers, Amazon-Photo, and Coauthor-CS \citep{shchur2018pitfalls}. All datasets are valid under long-tailed and imbalanced settings with GraphENS \citep{park2021graphens} and GraphSHA \citep{li2023graphsha}. Additionally, we evaluate our method's performance across various GNN backbones, including GraphSAGE \citep{hamilton2017inductive}, GCN \citep{kipf2016semi}, and GAT \citep{velivckovic2017graph}. We also experiment with different hyper-parameters to explore their influence. In summary, the key contributions are as follows:

\begin{itemize}
    \item We identified defects in GraphENS related to the randomness anomalous connectivity problem. Depending on the random seed, GraphENS may synthesize nodes that confuse the GNN, and the presence of these nodes significantly degrades the performance of the model. 
    
    \item PNS effectively avoids the randomness anomalous connectivity problem and ensures that model performance does not significantly degrade under unfavorable random seeds. With PNS, the model maintains stable performance.
    
    \item Throughout extensive experiments, we demonstrate that our method not only mitigates the effects of unfavorable random seeds but also achieves excellent performance across various datasets. It consistently outperforms baseline models, showcasing robust and stable results.
\end{itemize}

Following the introduction, we provide a comprehensive review of related work on graph imbalance class problems and graph neural networks in Section \ref{sec: related work}. In Section \ref{sec: preliminaries}, we formally define the problem and introduce the notation used throughout this paper. We also provide a brief overview of the fundamental concepts of Graph Neural Networks (GNNs). We provide a detailed introduction to the Randomness Anomalous Connectivity Problem (RACP) in Section \ref{sec: rac problem}, including a thorough explanation, theoretical analysis, and the results of the comparison of experimental performance when RACP occurs. In Section \ref{sec: methodologies}, we provide an overview of the key concepts and definitions related to PNS. Moreover, we conduct a theoretical analysis of how PNS makes performance better. In Section \ref{sec: experiment}, we conduct experiments to verify the performance of PNS and perform an in-depth analysis of PNS, including an ablation study, visualization, and an investigation into the influence of the hyperparameter $\rho$. Additionally, we analyze the complexity of PNS. We provide an in-depth analysis of RACP in section \ref{sec: In-depth analysis of RACP}, including reproducibility experiments, analysis, and an examination of the problem across various models. Finally, we conclude our work in section \ref{sec: Conclusion}.

\begin{figure*}
    \centering
    \subfigure[GCN]{
        \label{fig: GCN confusion compared}
        \includegraphics[scale = 0.48]{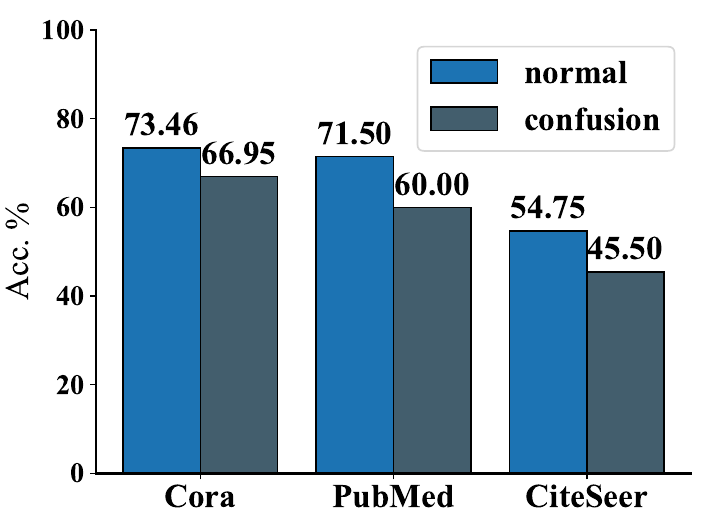}
    }
    \hspace{-6mm}
    \subfigure[GAT]{
        \label{fig: GAT confusion compared}
        \includegraphics[scale = 0.48]{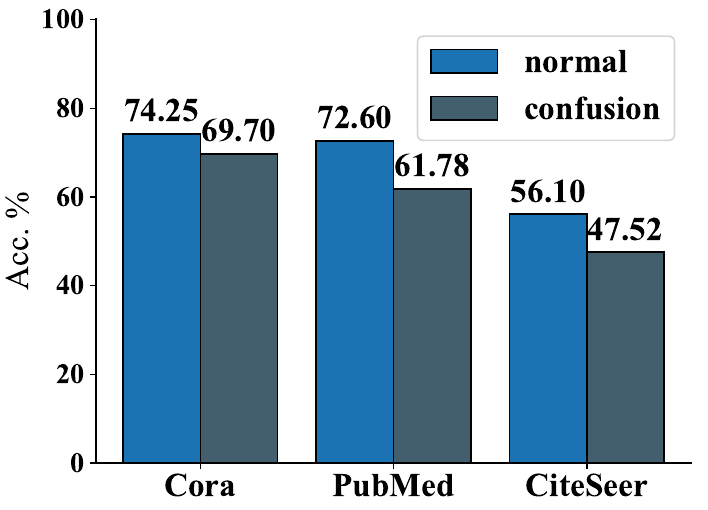}
    }
    \hspace{-6mm}
    \subfigure[GraphSAGE]{
        \label{fig: SAGE confusion compared}
        \includegraphics[scale = 0.48]{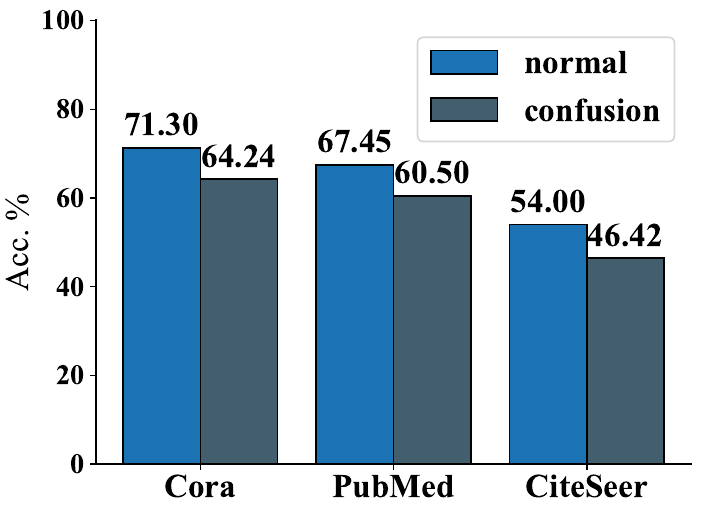}
    }
    \caption{The experiment resulted in a randomness anomalous connectivity problem in three different GNN backbones. We simulate an extreme scenario where we compare the performance of using only pure nodes versus only using confusion nodes for synthesis. In addition, we have repeated the experiment five times using different datasets and backbones to ensure the robustness of our results. If we only synthesize the confusion nodes, the model's performance has sharply declined. However, if we only utilize the pure node to synthesize, the model's performance remains good. In the following experiment, we can find that PNS not only tackles the randomness anomalous connectivity problem but also results in a better performance of the model.}
    \label{fig: confusion compare}
\end{figure*}

\section{Related Work} \label{sec: related work}

\subsection{GNN for Graph Imbalance Class Problem}

The imbalance class problem widely exists in the world \citep{japkowicz2002class, johnson2019survey}, like molecules' distribution in proteins. What is the imbalance class problem? The imbalance class problem can be summarized as an uneven distribution of the number of different classes in the training set. Models represented by machine learning and deep learning easily over-represent the majority class and vice versa. The imbalance class problem can cause terrible model performance \citep{song2022tam}. The graph imbalance class problem shares some similarities with the imbalance class problem, but there are some notable differences. The graph imbalance class problem is characterized by not only a disparity in quantity but also a topological structure\citep{park2021graphens, song2022tam}. In the realm of graph-based models, addressing the class imbalance problem can be approached in two main ways: node synthesis \citep{park2021graphens, qu2021imgagn, zhao2021graphsmote}, and loss modification \citep{chen2021topology, song2022tam, liu2021tail}. 

Node synthesis is the primary method in topological data, which synthesizes minor nodes to tackle the problem of quantity imbalance. GraphSMOTE \citep{zhao2021graphsmote} synthesizes the minor class by two minor nodes in the same class using SMOTE \citep{chawla2002smote}, also using an edge predictor for the synthesized nodes. GraphENS \citep{park2021graphens} considers the ego network centered around a minor node and combines another random node to synthesize new minor nodes. GraphSHA \citep{li2023graphsha} employs a novel approach to synthesizing minor nodes by leveraging difficult samples, effectively expanding the decision boundary of minor nodes. ImGAGN \citep{qu2021imgagn} and DR-GCN \citep{shi2020multi} utilize the generative adversarial network (GAN) \citep{goodfellow2020generative} to synthesize the minor nodes. For these methods, the synthesis of minor nodes often neglects the topological structure of the sample nodes. Furthermore, the sampling node with a confused topological structure can be challenging to process by GNN, as the message-passing mechanism tends to amplify the confusion.

The loss modification method reweights the weights of minor nodes in the training process. In terms of topological imbalance, the loss modification method solves the graph imbalance problem. TAM \citep{song2022tam} weights minor classes appropriately based on node label distribution. ReNode \citep{chen2021topology} adaptively adjusts the influence of labeled nodes based on their relative positions to class boundaries. Our method also considers the topology structure of the graph, but instead of modifying the loss function, we apply a filter operation to address the imbalance issue. Some other methods are also used to tackle the graph imbalance class problem, such as contrastive learning \citep{cui2023hybrid, zeng2023imgcl} or mathematical methods \citep{yan2024rethinking, zhu2023balanced}. 

Existing GNN imbalance methods typically address either the quantity imbalance, which arises from unequal class distributions, or the topology imbalance, which stems from differences in graph structure. However, a significant oversight in most existing methods is the neglect of the Randomness Anomalous Connectivity Problem (RACP) that arises during the node synthesis process, which can substantially compromise the stability and performance of GNN models. PNS is specifically designed to address RACP that arises during the node synthesis process, thereby not only mitigating the impact of random seeds but also yielding better performance.

\subsection{Graph Neural Network}

Graph neural network (GNN) is a type of neural network specifically designed to process graph-structured data. It leverages the connections between nodes (edges) to propagate and aggregate information, allowing it to learn representations for each node or the entire graph. These representations can be used for various tasks such as node classification, graph classification, link prediction, etc \citep{wu2020comprehensive, xu2018powerful}. The existing GNN methods can be briefly divided into two categories: spectral-based \citep{bruna2014spectral, tang2019chebnet} and spatial-based \citep{ atwood2016diffusion, duvenaud2015convolutional}.

Spectral-based GNN defines convolutional computation similar to the Fourier transform by computing the Laplace operator of the graph. However, the early computational complexity of this method was initially quite high, making it prohibitively expensive for many applications \citep{bruna2014spectral}. With the use of Chebyshev polynomials to approximate the results, computational efficiency has been improved. GCN \citep{kipf2016semi} made a significant approximation, retaining only the first two terms of the polynomial. The GCN is currently one of the most widely used GNNs.

Spatial-based GNNs are a class of GNN models that focus on the local structure and features of nodes and their neighbors \citep{gao2023survey}. Unlike spectral-based methods, spatial-based methods operate directly on the graph structure, using message-passing mechanisms to aggregate information from nodes and their neighbors. GraphSAGE \citep{hamilton2017inductive} samples a fixed-size set of neighbors for each node and uses an aggregation function to aggregate information from these neighbors. GAT \citep{velivckovic2017graph} introduces an attention mechanism to compute the importance weights of neighboring nodes, allowing for flexible aggregation of information. There are also many derived models \citep{lai2020policy, zhou2019meta}. PC-CONV \citep{li2024pc} represents one of the impressive GNNs.

However, traditional GNN models often neglect the issue of class imbalance inherent in graph data, which can lead to suboptimal performance when applied to imbalanced graph datasets. PNS uniquely addresses both the class imbalance issue in graph data and RACP in the node synthesis process, thereby effectively mitigating the problem of unstable performance that plagues current models.

\section{Preliminaries} \label{sec: preliminaries}

\subsection{Notation}

In this paper, our task focuses on a node classification task. The task is performed in an undirected graph $\mathcal{G}\, (\mathcal{V}, \mathcal{X}, \mathcal{A}, \mathcal{Y})$, where $\mathcal{V}$ = $\{v_1, v_2, ..., v_n\}$ is a set with $|\mathcal{V}|$ nodes. $\mathcal{X} \in R^{|\mathcal{V}| \times d}$ is a set of node features where $d$ denotes the dimension of $\mathcal{X}$, and $\mathcal{X}_i$ ($i$-th row of $\mathcal{X}$) denotes the feature of a node $v_i$. $\mathcal{A} \in R^{|\mathcal{V}| \times |\mathcal{V}|}$ is an adjacent matrix of $\mathcal{G}$. $\mathcal{Y}$ = $\{1,...,c\} $ is a class label set with $c$ class in total, where $\mathcal{Y}_i$ denotes the $i$-th class, $\mathcal{Y}(v_i)$ denotes the label of node $v_i$, and $v_{i, j}$ denotes node $v_i$ with label $j$. We use $\mathcal{N}(v_i)$ to denote the one-hop neighbors of node $v_i$. Moreover, we use $v_{Ai}$ to represent the node $v_i$ with the anomalous connectivity problem, $v_{\bar{A}i}$ on the contrary. 
We also use $\rho$ to represent the imbalance ratio between the most frequent class and the less frequent class, where $\rho$ = $\frac{|\{v \in \mathcal{V} | \mathcal{Y}(v) = \mathcal{Y}_{\textit{most freq}}\}|}{|\{v \in \mathcal{V} | \mathcal{Y}(v) = \mathcal{Y}_{least freq}\}|}$. $\mathcal{Y}_{\textit{most freq}}$ denotes the class with the highest frequency of occurrence, that is, the class with the largest quantity. $\mathcal{Y}_{\textit{least freq}}$ denotes the class with the least frequency of occurrence. Notice that the definitions of $\rho$ in GraphSMOTE \citep{zhao2021graphsmote} and GraphENS \citep{park2021graphens} are reciprocal of each other. Here, we follow the definition in GrapnENS.

\subsection{GNN for Node Classification}

GNN is excellent for handling graphs-based datasets and extracting the features hidden in the graph sharply. GNN can be divided into two main components: aggregation and update. The aggregation step involves gathering information from neighboring nodes, while the update step updates the representation of each node based on the aggregated information. For node $v$, the representation updating can be defined as:
\begin{equation}
    h_v^{(l)} = \text{UPDATE}\left( h_v^{(l-1)}, \text{AGG}\left( \{h_u^{(l-1)}:u \in \mathcal{N}(v)\} \right) \right),
\end{equation}
$h_v^{(l)}$ denotes the representation of node $v$ at l-th GNN layer. $h_v^{(l - 1)}$ is the previous layer representation of node $v$. \text{AGG} is a function that aggregates the information of node $v$ with its neighbors. \text{UPDATE} is a function that updates the representation of node $v$ by $h_v^{(l - 1)}$ with its aggregation information. Notice that $h_v^{(0)}$ = $\mathcal{X}$. After obtaining node embeddings, they are passed through a projector layer to yield the final classification output $\hat{y}_i$. Finally, Optimize the model using a loss function, such as cross-entropy loss, on the labeled data.
$$
\mathcal{L} = \frac{1}{|\mathcal{V}_{\text{train}}|} \sum_{i \in \mathcal{V}_{\text{train}}} \text{CrossEntropy}(\hat{y}_i, y_i).
$$

\subsection{Problem Definition}

The goal of the node synthesis step in addressing the class imbalance problem is to generate a balanced graph by synthesizing minority-class nodes, thereby mitigating the impact of the imbalance on the original graph $\mathcal{G}$. In this paper, we also reduce the selection boundaries of nodes to enable the model to synthesize more effective nodes for training. After data augmentation, we used the augmented graph $\mathcal{G}^{\prime}$ as input for the GNN. The GNN then performs a standard graph node classification task using the augmented graph $\mathcal{G}^{\prime}$.

\begin{figure}
    \centering
    \includegraphics[scale=0.41]{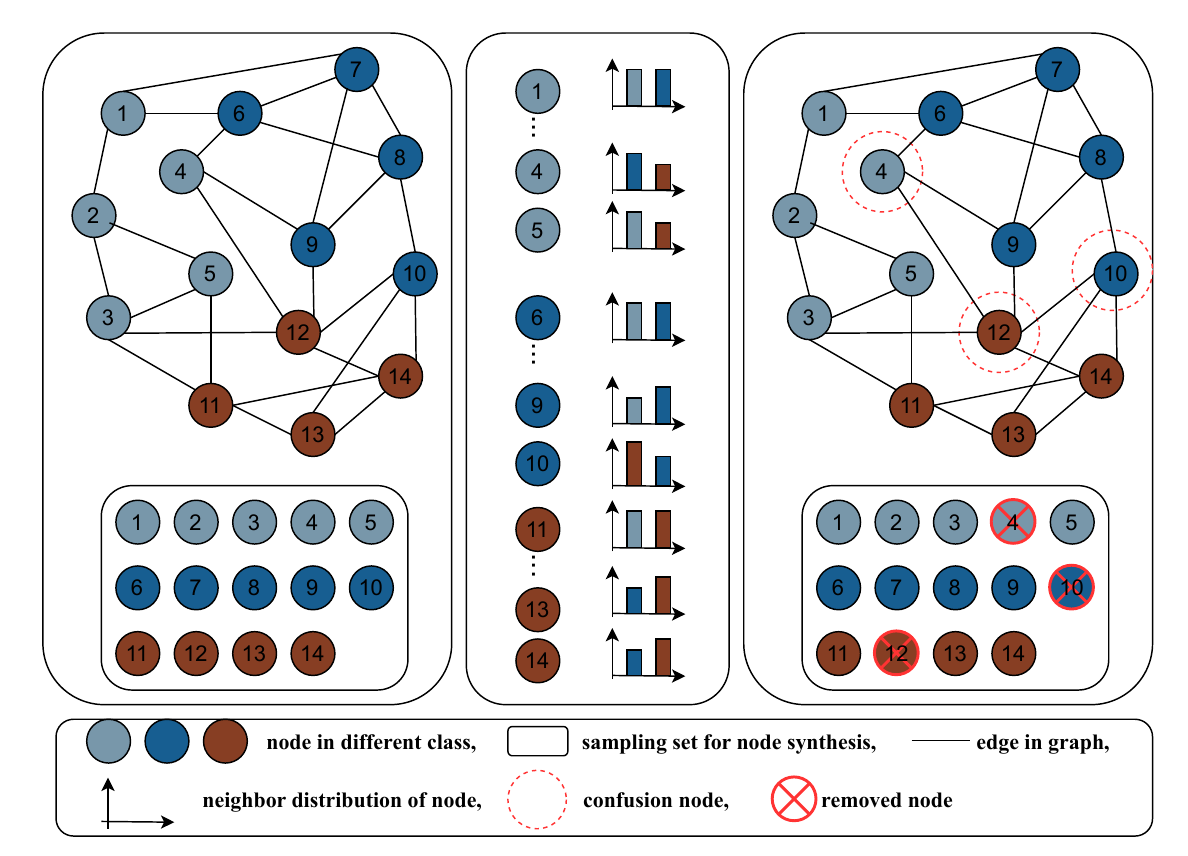}
    \caption{The toy handling process of PNS. The left shows the raw graph with its node sample boundary. Nodes 1-5 and 6-10 are the majority class, and nodes 11-14 are the minority class. PNS first calculates the neighbor label distribution of the training set, like in the middle. After that, PNS would distinguish the confusion node based on the purity of each node. Finally, PNS removes the confusion node from the sample set for pure node synthesis, like the nodes 4, 10, and 12 on the right, which reduces the sample boundary effectively.}
    \label{fig: pns progress}
\end{figure}

\section {Randomness Anomalous Connectivity Problem} \label{sec: rac problem}

The Randomness Anomalous Connectivity Problem (RACP) refers to the significant decrease in algorithmic performance caused by synthesizing nodes with different random seeds. While it may be difficult to fully resolve, one possible workaround is to change the random seed. We observed the presence of RACP in our experiments with GraphENS  \citep{park2021graphens} when  altering the random seed. The experiment result can be seen in Fig. \ref{fig: confusion compare}. Here, we analysis the cause of RACP. 

\subsection{Explaination of RACP}
\label{Explaination of RACP}

The RACP arises during node synthesis, where certain random seeds cause the model to preferentially select nodes with AC and use them for synthesis. Since many factors are influenced by random seeds, we hypothesize that node selection in synthesis-based methods for graph imbalance classification is heavily dependent on the random seeds. To address this, we introduce the concept of anomalous connectivity (AC), which refers to a node with an abnormal neighbor distribution that results in an excessive number of false positives during the quantity-based compensation process \citep{song2022tam}. Abnormal neighbor distribution refers to a situation where a node's neighbor distribution deviates from the typical connectivity pattern within its own class, like node 12 in Fig. \ref{fig: pns progress}. Then, we demonstrate our hypothesis through theory and experiments with the AC problem definition.

\subsection{Theoretical Analysis of RACP}
\label{sec: Theoretical Analysis of RACP}

We analyze RACP from a theoretical perspective, examining the underlying causes and factors that influence it.

We first define a node $v_A$ as follows:
\begin{equation}
\begin{split}
    & v_A  = \{v_{i,j} | \frac{|w|}{|\mathcal{N}(v_{i, j})|} < \rho \},  w \in \{\mathcal{N}(v_{i, j})\cap \mathcal{Y}(w) = j\}, \\
    & i \in [0, |\mathcal{V}|] \, \, \, and \, \, \,  j \in [0, c).
\end{split}
\end{equation}
$w$ is the neighbors of node $v_{i, j}$ whose class is $j$. $c$ is the total number of classes in a dataset. $v_A$ represents a node exhibiting the anomalous connectivity problem. It also means $P(\mathcal{Y}(\mathcal{N}(v_{i, j}))=\bar{j}) \gg P(\mathcal{Y}(\mathcal{N}(v_{i, j}))=j)$.
Based on the explanation in Section \ref{Explaination of RACP}, we can deduce that the node set $\mathcal{V}$ can be partitioned into two disjoint sets:  $\mathcal{V}_{\bar{A}}$ $\cup$ $\mathcal{V}_{A}$, namely $\mathcal{V}$ = $\mathcal{V}_{\bar{A}}$ $\cup$ $\mathcal{V}_{A}$. Specifically,  $\mathcal{V}_{A}$ consists of nodes with anomalous connectivity problems, while $\mathcal{V}_{\bar{A}}$ comprises nodes without such issues.      

Let $P(\mathcal{V}_A)$ = $p$ and $P(\mathcal{V}_{\bar{A}})$ = 1 - $p$, representing the probabilities of sampling a node with AC and a node without AC during the node synthesis process, respectively. Assuming the sampling process is performed with replacement, and our goal is to sample $n$ nodes, including $\alpha$ nodes with AC, we have:
\begin{equation} \label{equ: sampleing process}
    \begin{split}
        P(\mathcal{V}_{target}) & = P(Sample(\mathcal{V})) \\
                             & = P(Sample(\mathcal{V}_{\bar{A}}) \cup \mathcal{V}_{A}) \\
                             & =P( \{\mathcal{V}_{A_1},..., \mathcal{V}_{A_\alpha}, \mathcal{V}_{\bar{A_1}},..., \mathcal{V}_{\bar{A}_{n-\alpha}}\}) \\
                             & = \binom{n}{\alpha} p^\alpha(1 - p)^{n - \alpha}.
    \end{split}                 
\end{equation}

The entire sampling process can be modeled as a binomial distribution. In Equation \ref{equ: sampleing process}, $p$ is a constant, representing the fixed probability of sampling an AC node. Let $p_j$ denote the probability of sampling an AC node from class $j$, and $n_j$ denote the total number of nodes in class $j$. The aggregate probability $p$ can then be computed as:
\begin{equation}
    \begin{split}
        p & = \frac{\sum_{j=1}^{c} p_j n_j}{\sum_{i=1}^c n_j} \\
          & = \frac{\sum_{j=1}^{c} p_j n_j}{|V|} \\
          & = \frac{\sum_{j=1}^{c} \frac{\sum_{i=1}^{n_j} I(\frac{|w|}{|\mathcal{N}(v_{i, j})|} < \rho)}{n_j}n_j}{|V|}, \{Y(w) = j| w \in \mathcal{N}(v_{i,j})\}        \\    
          & = \frac{\sum_{j=1}^{c} \sum_{i=1}^{n_j} I(\frac{|w_j|}{|\mathcal{N}(v_{i,j})|} < \rho)}{|V|}.
    \end{split}
\end{equation}

Here $c$ is the total number of classes in a dataset. $I(\cdot)$ is am indicator function, $I(\cdot)$ = 1 if the condition is true, and $I(\cdot)$ = 0 otherwise.

We have an additional parameter $\alpha$ in Equation \ref{equ: sampleing process}, which is influenced by two factors: the probability $p$ and the random seed. $p$ is constant in an specific dataset. We note that PyTorch employs a Pseudo-Random Number Generator (PRNG) to generate random seeds, which in turn affects the sampling of node sequences through the sample function. As a result, the value of $\alpha$ is sensitive to the random seed used, leading to different values of $\alpha$ for different seeds. It means 
$$\mathcal{V}_{target} = \{\mathcal{V}_{A}^{s} \cup \mathcal{V}_{\bar{A}}^{s}\}, \ \ |V_{A}^{s}| = \alpha \in [0, |\mathcal{V}_{target}|]$$ 
to a specific random seed $s$. The use of different random seeds yields distinct node sequences for node synthesis, which is a primary contributor to RACP. Here, we revise the definition of RACP based on Equation \ref{equ: sampleing process}:
\begin{definition}
    \rm The randomness-induced anomalous connectivity problem (RACP) arises during stochastic node synthesis when a particular random seed causes the sampling process to select nodes exhibiting anomalous connectivity (AC) patterns. It can be expressed below: 
    \begin{equation}
        P^s(V_{target}) \approx \binom{n}{\alpha} (1 - p)^{n - \alpha}
    \end{equation}
    $ P^s(V_{target})$ denotes the target node synthesize distribution with a specific seed $s$. The whole equation means that with $s$ the model tends to select nodes with AC to synthesize. 
\end{definition}

Finally, we discuss how RACP degrades the performance of the model. We provide the following concise proof:

\begin{proof}
How RACP impair the model performance? We succinctly represent the message-passing process of GNNs as follows: 
\[
h^{l + 1}_{v,c} = \beta h^{l}_{\mathcal{N}(v), c} + (1 - \beta)h^{l}_{\mathcal{N}(v),\bar{c}},\ \  \beta \in (0, 1).
\]

 Assuming $\mathcal{Y}(v)$ = c, if $v \in \mathcal{V_A}$,  we have: 
 \[
 1 - \beta \gg \beta,
 \]

results in: 
 \[
  ||h_{v,c} - h_{c}||  \gg ||h_{v, c} - h_{ \bar{c}}|,
 \]

 so that in loss calculation, we have: 
 \[
 \mathcal{Y}(v)log(\hat{\mathcal{Y}}(v) = \bar{c}) \gg \mathcal{Y}(v)log(\hat{\mathcal{Y}}(v) = c),
 \]

 which results in:
\[
\mathbf{E}[\hat{Y}(v) = \bar{c}] \gg \mathbf{E}[\hat{Y}(v) = c],
\] when $\mathcal{Y}(v) = c$.

\end{proof}

\begin{algorithm}[t]
    \renewcommand{\algorithmicrequire}{\textbf{Input:}}
    \renewcommand{\algorithmicensure}{\textbf{Output:}}
    \caption{Pure node sampling algorithm}
    \label{algo: pns}
    \begin{algorithmic}[1]
        \REQUIRE Graph $\mathcal{G}\, (\mathcal{V}, \mathcal{X}, \mathcal{A}, \mathcal{Y})$, Confusion rate $r$
        \ENSURE $prue\_\mathcal{V}_{train}$, the pure training set which filters out the confusion nodes.
        
    \STATE $\mathcal{V}_{train}$ = train\_set($\mathcal{V}$)

        \STATE \texttt{// statistic the neighbor label distribution $\mathcal{D}$; $c_i$ denotes the class i.}
        \FOR{$\mathcal{V}_{c_i}$ in $\mathcal{V}_{train}$} 
            \FOR{$v_{c_i}$ in $\mathcal{V}_{c_i}$}
                \STATE $\mathcal{D}_{c_i}$ = $\mathcal{D}_{c_i} \cup \mathcal{Y}(\mathcal{N}(v_{c_i}))$
            \ENDFOR
            \STATE $\mathcal{D} = \mathcal{D} \cup \mathcal{D}_{c_i}$
        \ENDFOR

        \STATE \texttt{// calculate the $\mathcal{P}_i$ of each node.}
        \FOR{$\mathcal{D}_{c_i}$ in $\mathcal{D}$}
            \IF{$len(\mathcal{D}_i) != 0$}
                \STATE $P_{c_i} = \frac{\mathcal{D}_{c_i}.count(c_i)}{len(\mathcal{D}_i)}$
            \ENDIF
            \STATE $P = P \cup P_{c_i}$
        \ENDFOR
        
        \FOR{$v_{c_i}$ in $\mathcal{V}_{train}$}
            \IF{$\mathcal{P}_{c_i}(v_{c_i}) > r$}
                \STATE $prue\_\mathcal{V}_{train}$ = $prue\_\mathcal{V}_{train} \cup v_{c_i}$
            \ENDIF
        \ENDFOR

        \STATE \textbf{return} $prue\_\mathcal{V}_{train}$
        
    \end{algorithmic}

\end{algorithm}

\section{Methodologies} 
\label{sec: methodologies}
We introduce our method in this section. We conducted a reverse analysis based on the experiment in the previous section. By selecting only pure nodes, we can reduce the boundaries of node selection and mitigate the risk of poor performance caused by random seed selection of confusing nodes. PNS not only reduces the impact of confusing nodes during message passing but also addresses the random anomalous connectivity problem. We use node purity to evaluate whether a node is confusing, controlled by the hyperparameter confusion rate $r$. Furthermore, we analyze the computational complexity and describe our method in Algorithm \ref{algo: pns}. Noticed that PNS is applicable to transductive settings and is publicly available at https://github.com/flzeng1/PNS.

\subsection{Node Purity Recognition}

We propose two key definitions used in our algorithm to estimate whether a node is a confusion node: neighbor label distribution $\mathcal{D}$ and node purity $\mathcal{P}$.

\begin{definition} 
  \rm  Neighbor labels distribution (NLD), $\mathcal{D} \in \mathcal{R}^{|\mathcal{V}| \times c}$, in this paper, only calculates the label quantity of each node and is defined as below. $\mathcal{D}_i$ denotes the set of neighbor labels distribution of $v_i$, and $\mathcal{D}_{i, j}$ denotes the element in $\mathcal{D}_i$ where class = $j$.
\begin{equation}
    \mathcal{D}_i =\bigcup_{j \in \mathcal{Y}} \{ {v \in \mathcal{N}(v_i)\,|\, j = \mathcal{Y}(v)}\}.
\end{equation}
\end{definition}

\begin{definition} 
  \rm  Node purity $\mathcal{P}$ is a metric that is used to determine whether a node is ``pure" enough for embedding. Let $\mathcal{P}_i$ denote the purity of node $v_i$, then $\mathcal{P}$ is defined as:
\begin{equation}
    \mathcal{P}_i = \frac{1}{\sum_{j \in \mathcal{Y}} |\mathcal{D}_{i,j}|} |\mathcal{D}_{i, \mathcal{Y}(v_i)}|.
\end{equation}
\end{definition}

Node purity recognition is a method to calculate the purity of each node. The goal of node purity recognition is to distinguish the confusion rate of each node. Firstly, we should calculate the neighbor label distribution $\mathcal{D}_i$ of each node based on $\mathcal{N}$. Then, we calculate the node purity $\mathcal{P}_i$ to ensure whether a node would confuse the GNN.

Notice that the definition of \textit{NLD}'s name is the same as TAM \citep{song2022tam} and comes from Geom-GCN \citep{pei2019geom}, but different in computation. We define these two terms since we need the label distribution to determine the pure nodes. In supervised learning, we directly utilize the label information. 

\subsection{Pure Node Sampling}

With the node purity set $\mathcal{P}$, we can reduce the class node boundary. We use the purity of each node $\mathcal{P}_i$ compared with the hyperparameter confusion rate $r$ to determine which node would be removed from the class node sampling boundaries $\mathcal{C}$. The compared method is defined as follows: 
$$ \mathcal{C} = \left\{
\begin{aligned}
& \bigcup \varnothing, \, \mathcal{P}_i < r \\
& \bigcup v_i, \, \mathcal{P}_i >= r .\\
\end{aligned}
\right.
$$

It should be noted that the core idea of PNS shares conceptual similarities with traditional noise filtering and sample selection methods. However, a key distinction is that PNS filters nodes based on NLD. This enables the detection of nodes that may confuse the learning ability of GNN models in imbalanced scenarios. Furthermore, a detailed theoretical analysis of PNS is provided in Subsection \ref{sec: Theoretical Analysis of PNS}, which offers insights into the underlying mechanisms that drive its effectiveness.

\subsection{Theoretical Analysis of PNS}
\label{sec: Theoretical Analysis of PNS}

As discussed in Section \ref{sec: rac problem}, we have established that nodes with anomalous connectivity problems can compromise both model performance and the node synthesis process. This part of the paper presents a theoretical examination of the ways in which PNS mitigates these issues from a conceptual perspective.

Upon applying PNS, the nodes used for synthesis and training are drawn from a new node set, denoted as PNS($\mathcal{V}$) $\subset \mathcal{V}$, which has been filtered to exclude a certain number of nodes with anomalous connectivity (AC). Consequently, PNS can mitigate the impact of varying random seeds. Subsequently, in the node synthesis process, the target node set $\mathcal{V}_{target}$  is obtained by sampling from the filtered node set PNS($\mathcal{V}$), yielding $\mathcal{V}_{target}$ = Sample(PNS($\mathcal{V}$)) = $\{v_{\bar{A}_1}, ... ,v_{\bar{A}_n} \}$.

\textbf{How does PNS improve model performance?} Here, we demonstrate how PNS improves model performance by providing a theoretical analysis:

\begin{proof} \label{proof: PNS make model performance better}
   How does PNS make model performance better? We succinctly represent the message-passing process of GNNs as follows:
   \[
   h^{l + 1}_{v,c} = \beta h^{l}_{\mathcal{N}(v), c} + (1 - \beta)h^{l}_{\mathcal{N}(v),\bar{c}},\ \  \beta \in (0, 1).
   \]
   Assuming $\mathcal{Y}(v)$ = c, if $v \in$ PNS $ (\mathcal{V})$, we have:
   \[
   \beta \gg 1 - \beta,
   \]
   results in:
   \[
   ||h_{v, c} - h_{ \bar{c}}|| \gg ||h_{v,c} - h_{c}||,
   \]
   so that in loss calculation, we have:
   \[
   \mathcal{Y}(v)\log(\hat{\mathcal{Y}}(v) = c) \gg
   \mathcal{Y}(v)\log(\hat{\mathcal{Y}}(v) = \bar{c}),  
   \]
   which results in:
   \[
   \mathbf{E}[\hat{Y}(v) = c] \gg \mathbf{E}[\hat{Y}(v) = \bar{c}], 
   \]
   when $\mathcal{Y}(v) = c$.
\end{proof}

\textbf{How does PNS impact label propagation?} PNS employs purity to quantify the anomalous connectivity (AC) degree of each node. Nodes are subsequently eliminated by comparing their purity against the confusion rate $r$. Furthermore, PNS decouples the random seed from the node synthesis process. For label propagation, we have:
\begin{equation*}
    h_v^{(l)} = \texttt{UPDATE}\left(h_v^{(l-1)},\ \texttt{AGG}\left(\left\{ h_u^{(l-1)} \mid u \in \mathcal{N}(v) \right\}\right)\right).
\end{equation*}

AC nodes can obscure the learned representations, as node embeddings are generally influenced by the features of their neighbors \citep{kipf2016semi}. Additionally, AC nodes exhibit anomalous connectivity patterns \citep{song2022tam}, which distort this process.  The anomalous connectivity pattern interferes with the  $\texttt{AGG}\left(\left\{ h_u^{(l-1)} \mid u \in \mathcal{N}(v) \right\}\right)$ during the \texttt{UPDATE}. As a result, the aggregated representation tends to align with features from other classes, which means $\mathbf{E}_{v \sim V}[h_{v}^{(l)}| f (h_{v}^{(l)})] \to \mathbf{E}_{z \in V_{A}}[h_{z}^{(l)}|f(h_{z}^{(l)}) \neq c, \mathcal{Y}(z) = c]$, thereby introduces confusion to GNN $f$. In the node synthesis stage, the sampled node set inevitably includes AC nodes, where $V_{sample}$ = $\{V_A \cup V_{\bar{A}}\}$. If RACP happens, $V_{target}$ $\to$ $\{V_{A}\}$. Using AC nodes for synthesis interferes with the $\texttt{AGG}$ process, causing the synthesized nodes to represent other classes, $\mathbf{E}_{v \sim V_{target}^s}[h_{v}^{(l)}|f(h_{v}^{(l)})] \to \mathbf{E}_{z \sim V_{A}}[h_{z}^{(l)}|f(h_{z}^{(l)}) \neq c, \mathcal{Y}(z) = c]$. $s$ denotes a specific seed, $z_A$ denotes a node with AC issue.

 In the label propagation process, PNS employs purity to quantify the AC degree of each node. By comparing the node purity against the confusion rate $r$, PNS selectively excludes a proportion of AC nodes from both node synthesis and subsequent training. The functionality of PNS can be formally expressed as the construction of a refined training node set, $V^\prime$ = $\{u \in V \cap \mathcal{P}_u \ge r\}$. PNS mitigates the adverse influence of anomalous neighbors on the central node, enabling $\texttt{AGG}\left(\left\{ h_u^{(l-1)} \mid u \in \mathcal{N}(v) \right\}\right)$ to produce a more reliable representation. Consequently, with PNS, a node’s representation is more likely to align with its true class, $\mathbf{E}_{v \sim V_{target}^s}[h_{v}^{(l)}|f(h_{v}^{(l)})] \to \mathbf{E}_{z \sim V^\prime}[h_{z}^{(l)}|f(h_{z}^{(l)}) \to c, \mathcal{Y}(z) = c]$.

\section{Experiments} \label{sec: experiment}

\begin{table}[b]
\caption{Data details.}
\label{table: dataset deatils}
\begin{tabular}{l|cccc}
\hline
\multicolumn{1}{c|}{Datasets} & Nodes & Edges  & Features & Classes \\ \hline
Cora                          & 2078  & 10556  & 1433     & 7       \\
CiteSeer                      & 3327  & 9104   & 3703     & 6       \\
PubMed                        & 19717 & 88648  & 500      & 3       \\
Photo                         & 7650  & 119081 & 745      & 8       \\
Computer                      & 13752 & 245861 & 767      & 10      \\
CS                            & 18333 & 81894  & 6805     & 15      \\ \hline
\end{tabular}

\end{table}

\begin{table*}[h]
    \footnotesize
    \setlength{\tabcolsep}{3pt}
    \renewcommand\arraystretch{1.3}
    \caption{Imbalance node classification results of PNS compared with other baselines. We set the imbalance ratio in an extreme setting ($\rho$ = 100) with standard errors five times.}
\label{table: citation networks}

\begin{tabular}{ll|lll|lll|lll}
\hline
\multirow{2}{*}{\textbf{}} &
  \textbf{Dataset} &
  \multicolumn{3}{c|}{\textbf{Cora-LT}} &
  \multicolumn{3}{c|}{\textbf{CiteSeer-LT}} &
  \multicolumn{3}{c}{\textbf{PubMed-LT}} \\ \cline{2-11} 
 &
  $\rho$=100 &
  \multicolumn{1}{c}{Acc.} &
  \multicolumn{1}{c}{bAcc.} &
  \multicolumn{1}{c|}{F1} &
  \multicolumn{1}{c}{Acc.} &
  \multicolumn{1}{c}{bAcc.} &
  \multicolumn{1}{c|}{F1} &
  \multicolumn{1}{c}{Acc.} &
  \multicolumn{1}{c}{bAcc.} &
  \multicolumn{1}{c}{F1} \\ \hline
\multicolumn{1}{c}{\multirow{9}{*}{\rotatebox{90}{GCN}}} &
  Vanilla &
  73.60\scriptsize$\pm$0.43 &
  64.13\scriptsize$\pm$0.71 &
  63.78\scriptsize$\pm$0.68 &
  54.56\scriptsize$\pm$0.41 &
  47.79\scriptsize$\pm$0.38 &
  43.17\scriptsize$\pm$0.57 &
  69.48\scriptsize$\pm$0.98 &
  56.54\scriptsize$\pm$0.79 &
  51.06\scriptsize$\pm$0.73 \\ \cline{2-11} 
\multicolumn{1}{c}{} &
  Reweight &
  72.38\scriptsize$\pm$0.18 &
  64.86\scriptsize$\pm$0.28&
  65.23\scriptsize$\pm$0.29 &
  54.60\scriptsize$\pm$0.71 &
  48.09\scriptsize$\pm$0.81 &
  43.67\scriptsize$\pm$1.17 &
  67.90\scriptsize$\pm$0.73 &
  55.26\scriptsize$\pm$0.58 &
  49.92\scriptsize$\pm$0.50 \\
\multicolumn{1}{c}{} &
  Upsampling &
  71.64\scriptsize$\pm$0.49 &
  63.63\scriptsize$\pm$0.68 &
  63.97\scriptsize$\pm$0.85 &
  54.62\scriptsize$\pm$0.66 &
  48.03\scriptsize$\pm$0.68 &
  43.53\scriptsize$\pm$1.01 &
  65.76\scriptsize$\pm$0.15 &
  53.54\scriptsize$\pm$0.12 &
  48.26\scriptsize$\pm$0.11 \\
  \multicolumn{1}{c}{} &
  M-ILBO &
19.96\scriptsize$\pm$6.96 &
17.45\scriptsize$\pm$3.16 &
7.53\scriptsize$\pm$4.24 &
32.02\scriptsize$\pm$9.93 &
31.88\scriptsize$\pm$6.21 &
20.44\scriptsize$\pm$7.37 &
32.16\scriptsize$\pm$5.78 &
33.77\scriptsize$\pm$0.24 &
16.61\scriptsize$\pm$2.66 \\
  
\multicolumn{1}{c}{} &
  GraphSMOTE &
  71.48\scriptsize$\pm$0.21 &
  62.94\scriptsize$\pm$0.22 &
  63.39\scriptsize$\pm$0.17 &
  53.80\scriptsize$\pm$0.17 &
  47.25\scriptsize$\pm$0.19 &
  42.91\scriptsize$\pm$0.37   &
  67.90\scriptsize$\pm$0.24&
  55.84\scriptsize$\pm$0.18 &
  51.66\scriptsize$\pm$0.42 \\
\multicolumn{1}{c}{} &
  GraphENS &
  76.64\scriptsize$\pm$0.40 &
  71.29\scriptsize$\pm$0.69 &
  71.01\scriptsize$\pm$0.67 &
  62.22\scriptsize$\pm$0.48 &
  56.16\scriptsize$\pm$0.59 &
  55.09\scriptsize$\pm$0.72 &
  76.90\scriptsize$\pm$0.30 &
  71.11\scriptsize$\pm$1.23 &
  72.07\scriptsize$\pm$1.14 \\
\multicolumn{1}{c}{} &
  PC Softmax &
  78.16\scriptsize$\pm$0.63 &
  76.81\scriptsize$\pm$0.52 &
  75.06\scriptsize$\pm$0.58 &
  58.74\scriptsize$\pm$0.34 &
  56.40\scriptsize$\pm$0.22 &
  \textbf{57.36\scriptsize$\pm$0.16} &
  47.56\scriptsize$\pm$1.15 &
  54.16\scriptsize$\pm$0.96 &
  47.95\scriptsize$\pm$1.01 \\
\multicolumn{1}{c}{} &
   BAT &
75.50\scriptsize$\pm$0.22 &
66.54\scriptsize$\pm$0.38 &
66.38\scriptsize$\pm$0.75 &
59.26\scriptsize$\pm$0.52 &
52.56\scriptsize$\pm$0.49 &
49.07\scriptsize$\pm$0.67 &
67.50\scriptsize$\pm$0.46 &
55.03\scriptsize$\pm$0.35 &
49.90\scriptsize$\pm$0.30 \\
 &
   GNNCL &
70.88 \scriptsize$\pm$0.32 &
60.20\scriptsize$ \pm $0.57 &
58.13\scriptsize$ \pm $0.70 &

52.70\scriptsize$\pm$0.21 &
46.14\scriptsize$\pm$0.21 &
40.96\scriptsize$\pm$0.35 &

43.24\scriptsize$\pm$2.42 &
35.31\scriptsize$\pm$1.89 &
23.77\scriptsize$\pm$3.20 \\
 &

  GraphSHA &
  \textbf{79.36\scriptsize$\pm$0.29} &
  \textbf{73.87\scriptsize$\pm$0.38}&
  \textbf{75.28\scriptsize$\pm$0.30}&
  61.50\scriptsize$\pm$0.40 &
  56.18\scriptsize$\pm$0.34 &
  55.18\scriptsize$\pm$0.49 &
  71.24\scriptsize$\pm$0.50 &
  63.04\scriptsize$\pm$1.32 &
  61.98\scriptsize$\pm$1.27\\ \cline{2-11} 
\multicolumn{1}{c}{} &

  \textbf{PNS + BAT} &
75.74\scriptsize$\pm$0.45 &
66.88\scriptsize$\pm$0.72 &
67.10\scriptsize$\pm$0.95 &
59.32\scriptsize$\pm$0.37 &
52.61\scriptsize$\pm$0.37 &
49.19\scriptsize$\pm$0.54 &
68.22\scriptsize$\pm$0.38 &
55.63\scriptsize$\pm$0.32 &
50.50\scriptsize$\pm$0.39
  \\ 
\multicolumn{1}{c}{} &

  \textbf{PNS + ENS} &
  77.58\scriptsize$\pm$0.33 &
  72.37\scriptsize$\pm$0.54 &
  71.93\scriptsize$\pm$0.58 &
  \textbf{63.26\scriptsize$\pm$0.45} &
  \textbf{57.00\scriptsize$\pm$0.39} &
  54.87\scriptsize$\pm$0.59 &
  \textbf{79.94\scriptsize$\pm$0.58} &
  \textbf{76.03\scriptsize$\pm$1.06} &
  \textbf{76.73\scriptsize$\pm$0.89} \\ \hline
\multirow{9}{*}{\rotatebox{90}{GAT}} &
  Vanilla &
  73.54\scriptsize$\pm$0.33 &
  64.44\scriptsize$\pm$0.53 &
  64.06\scriptsize$\pm$0.77 &
  56.80\scriptsize$\pm$0.32 &
  50.23\scriptsize$\pm$0.29 &
  46.65\scriptsize$\pm$0.35 &
  70.55\scriptsize$\pm$0.25 &
  57.40\scriptsize$\pm$0.20 &
  51.92\scriptsize$\pm$0.18 \\ \cline{2-11} 
 &
  Reweight &
   73.62\scriptsize$\pm$0.79&
   66.19\scriptsize$\pm$1.19&
   66.54\scriptsize$\pm$1.37&
   53.92\scriptsize$\pm$0.22&
   47.40\scriptsize$\pm$0.22&
   41.91\scriptsize$\pm$0.43&
   63.16\scriptsize$\pm$0.42&
   51.45\scriptsize$\pm$0.34&
   46.28\scriptsize$\pm$0.34\\
 &
  Upsampling &
   72.94\scriptsize$\pm$0.67&
   65.90\scriptsize$\pm$0.96&
   66.19\scriptsize$\pm$1.09&
   54.46\scriptsize$\pm$0.20&
   47.92\scriptsize$\pm$0.19&
   42.78\scriptsize$\pm$0.29&
   64.10\scriptsize$\pm$0.28&
   52.21\scriptsize$\pm$0.23&
   47.04\scriptsize$\pm$0.23\\
    &
  M-ILBO &
53.14\scriptsize$\pm$0.64 &
36.78\scriptsize$\pm$0.64 &
34.11\scriptsize$\pm$0.81 &
47.04\scriptsize$\pm$0.44 &
41.05\scriptsize$\pm$0.39 &
32.28\scriptsize$\pm$0.48 &
44.42\scriptsize$\pm$0.72 &
36.39\scriptsize$\pm$0.62 &
25.83\scriptsize$\pm$1.19 \\
 &
  GraphSMOTE &
  73.76\scriptsize$\pm$0.51 &
  66.23\scriptsize$\pm$0.63 &
  67.07\scriptsize$\pm$0.92 &
  54.90\scriptsize$\pm$0.16&
  48.23\scriptsize$\pm$0.16 &
  43.78\scriptsize$\pm$0.32 &
  67.82\scriptsize$\pm$0.43 &
  55.91\scriptsize$\pm$0.52 &
  51.96\scriptsize$\pm$0.85 \\
 &
  GraphENS &
  77.52\scriptsize$\pm$0.21 &
  72.18\scriptsize$\pm$0.44 &
  71.98\scriptsize$\pm$0.65 &
  63.76\scriptsize$\pm$0.15 &
  57.62\scriptsize$\pm$0.22 &
  56.27\scriptsize$\pm$0.33 &
  76.10\scriptsize$\pm$0.10 &
  69.88\scriptsize$\pm$0.39 &
  70.77\scriptsize$\pm$0.44 \\
 &
  PC Softmax &
  73.38\scriptsize$\pm$1.07 &
  66.81\scriptsize$\pm$1.34 &
  65.76\scriptsize$\pm$1.52 &
  64.32\scriptsize$\pm$0.56 &
  61.41\scriptsize$\pm$0.69 &
  61.59\scriptsize$\pm$0.63 &
  77.26\scriptsize$\pm$1.02 &
  75.61\scriptsize$\pm$0.77 &
  74.63\scriptsize$\pm$0.91 \\
 &
    BAT &
76.84\scriptsize$\pm$0.22 &
68.53\scriptsize$\pm$0.28 &
67.16\scriptsize$\pm$0.42 &
59.30\scriptsize$\pm$0.58 &
52.49\scriptsize$\pm$0.58 &
49.09\scriptsize$\pm$0.73 &
65.80\scriptsize$\pm$0.69 &
53.57\scriptsize$\pm$0.56 &
48.44\scriptsize$\pm$0.55 \\

 &
GNNCL &
71.84\scriptsize$\pm$0.36 &
61.68\scriptsize$\pm$0.37 &
59.76\scriptsize$\pm$0.37 & 
54.82\scriptsize$\pm$0.12 & 
48.16\scriptsize$\pm$0.10 & 
43.55\scriptsize$\pm$0.16 & 
63.28\scriptsize$\pm$0.49 & 
51.59\scriptsize$\pm$0.43 &  
46.47\scriptsize$\pm$0.51 \\

 &
 
  GraphSHA &
  78.18\scriptsize$\pm$0.28 &
  72.67\scriptsize$\pm$0.35 &
  \textbf{74.06\scriptsize$\pm$0.40} &
  61.26\scriptsize$\pm$0.41 &
  56.05\scriptsize$\pm$0.23 &
  54.98\scriptsize$\pm$0.63 &
  68.82\scriptsize$\pm$0.29 &
  65.05\scriptsize$\pm$1.06 &
  64.26\scriptsize$\pm$0.95 \\ \cline{2-11}
 &

  \textbf{PNS + BAT} &
77.30\scriptsize$\pm$0.55 &
68.86\scriptsize$\pm$0.70 &
67.48\scriptsize$\pm$0.74 &
60.34\scriptsize$\pm$0.73 &
53.47\scriptsize$\pm$0.74 &
50.07\scriptsize$\pm$0.94 &
66.02\scriptsize$\pm$1.03 &
53.75\scriptsize$\pm$0.83 &
48.57\scriptsize$\pm$0.80 \\

\multicolumn{1}{c}{} &
 
  \textbf{PNS + ENS} &
  \textbf{78.20\scriptsize$\pm$0.24} &
  \textbf{73.07\scriptsize$\pm$0.24} &
  72.58\scriptsize$\pm$0.22 &
  \textbf{66.18\scriptsize$\pm$0.46} &
  \textbf{59.76\scriptsize$\pm$0.45} &
  \textbf{57.84\scriptsize$\pm$0.46}&
  \textbf{79.66\scriptsize$\pm$0.51 }&
  \textbf{77.17\scriptsize$\pm$1.03} &
  \textbf{77.42\scriptsize$\pm$0.72} \\ \hline
\multirow{9}{*}{\rotatebox{90}{SAGE}} &
  Vanilla &
  71.56\scriptsize$\pm$0.29 &
  60.27\scriptsize$\pm$0.36 &
  60.57\scriptsize$\pm$0.29 &
  50.30\scriptsize$\pm$0.40 &
  44.00\scriptsize$\pm$0.35 &
  39.34\scriptsize$\pm$0.55 &
  63.50\scriptsize$\pm$1.10 &
  51.72\scriptsize$\pm$0.88 &
  46.45\scriptsize$\pm$0.92 \\ \cline{2-11} 
\multicolumn{1}{c}{} &
  Reweight &
  72.66\scriptsize$\pm$0.57 &
  63.23\scriptsize$\pm$0.85&
  63.77\scriptsize$\pm$0.86 &
  54.88\scriptsize$\pm$0.33 &
  48.26\scriptsize$\pm$0.32 &
  43.47\scriptsize$\pm$0.34 &
  69.80\scriptsize$\pm$1.64 &
  57.79\scriptsize$\pm$1.63 &
  54.10\scriptsize$\pm$2.07 \\
\multicolumn{1}{c}{} 
 &
  Upsampling &
  71.62\scriptsize$\pm$0.66 &
  62.24\scriptsize$\pm$0.85 &
  62.76\scriptsize$\pm$0.95 &
  54.98\scriptsize$\pm$0.25 &
  48.44\scriptsize$\pm$0.17 &
  44.04\scriptsize$\pm$0.19 &
  73.44\scriptsize$\pm$0.93 &
  63.20\scriptsize$\pm$1.72 &
  62.04\scriptsize$\pm$2.31 \\
 &
   M-ILBO &
13.00\scriptsize$\pm$0.00&
14.29\scriptsize$\pm$0.00&
3.29\scriptsize$\pm$0.00&
7.70\scriptsize$\pm$0.00&
16.67\scriptsize$\pm$0.00&
2.38\scriptsize$\pm$0.00&
18.00\scriptsize$\pm$0.00&
33.33\scriptsize$\pm$0.00&
10.17\scriptsize$\pm$0.00 \\

  &
  GraphSMOTE &
  73.32\scriptsize$\pm$0.84&
  64.11\scriptsize$\pm$0.96 &
  64.86\scriptsize$\pm$0.91 &
  54.02\scriptsize$\pm$0.35 &
  47.78\scriptsize$\pm$0.37 &
  44.56\scriptsize$\pm$0.55 &
  73.60\scriptsize$\pm$1.46&
  63.85\scriptsize$\pm$1.85 &
  63.66\scriptsize$\pm$2.41 \\
 &
  GraphENS &
  76.46\scriptsize$\pm$0.29 &
  70.30\scriptsize$\pm$0.37 &
  69.83\scriptsize$\pm$0.39 &
  63.58\scriptsize$\pm$0.33 &
  57.12\scriptsize$\pm$0.32 &
  55.21\scriptsize$\pm$0.19 &
  77.80\scriptsize$\pm$1.20 &
  71.21\scriptsize$\pm$2.06 &
  72.35\scriptsize$\pm$2.20 \\
 &
  PC Softmax &
  \textbf{79.94\scriptsize$\pm$0.29} &
  \textbf{75.19\scriptsize$\pm$0.48} &
  \textbf{75.52\scriptsize$\pm$0.43} &
  \textbf{66.20\scriptsize$\pm$0.26} &
  \textbf{62.63\scriptsize$\pm$0.23} &
  \textbf{62.86\scriptsize$\pm$0.18} &
  80.02\scriptsize$\pm$0.06 &
  \textbf{78.95\scriptsize$\pm$0.11} &
  78.42\scriptsize$\pm$0.04 \\
 &
   BAT &
77.06\scriptsize$\pm$0.26 &
69.39\scriptsize$\pm$0.20 &
70.05\scriptsize$\pm$0.31 &
58.16\scriptsize$\pm$0.64 &
52.64\scriptsize$\pm$0.66 &
49.04\scriptsize$\pm$0.89 &
70.94\scriptsize$\pm$0.54 &
57.92\scriptsize$\pm$0.60 &
52.80\scriptsize$\pm$0.88 \\

 &

 GNNCL &
70.12\scriptsize$\pm$0.38 & 
59.73\scriptsize$\pm$0.70 &  
59.64\scriptsize$\pm$0.80 &
52.24\scriptsize$\pm$0.45 & 
45.87\scriptsize$\pm$0.40 &
40.73\scriptsize$\pm$0.50 &
65.62\scriptsize$\pm$0.16 &
53.43\scriptsize$\pm$0.13 &
48.17\scriptsize$\pm$0.13 \\

 &
 
  GraphSHA &
  78.60\scriptsize$\pm$0.18 &
  73.20\scriptsize$\pm$0.29 &
  74.24\scriptsize$\pm$0.23 &
  60.38\scriptsize$\pm$0.69 &
  56.90\scriptsize$\pm$0.60 &
  56.73\scriptsize$\pm$0.56 &
  66.32\scriptsize$\pm$0.50 &
  63.18\scriptsize$\pm$1.29 &
  61.64\scriptsize$\pm$1.08\\ \cline{2-11}
 &
   \textbf{PNS + BAT} &
78.56\scriptsize$\pm$0.53 &
72.04\scriptsize$\pm$0.89 &
72.65\scriptsize$\pm$0.89 &
58.58\scriptsize$\pm$0.39 &
52.97\scriptsize$\pm$0.33 &
49.27\scriptsize$\pm$0.54 &
71.16\scriptsize$\pm$0.66 &
58.18\scriptsize$\pm$0.65 &
53.21\scriptsize$\pm$0.84 \\

\multicolumn{1}{c}{} &

  \textbf{PNS + ENS} &
  76.72\scriptsize$\pm$0.25 &
  70.31\scriptsize$\pm$0.17 &
  69.87\scriptsize$\pm$0.26 &
  64.76\scriptsize$\pm$0.37 &
  58.45\scriptsize$\pm$0.36 &
  55.80\scriptsize$\pm$0.25 &
 \textbf{81.22\scriptsize$\pm$0.22} &
  78.93\scriptsize$\pm$0.15 &
  \textbf{79.15\scriptsize$\pm$0.11} \\ \hline
\end{tabular}

\end{table*}

\begin{table*}[h]
\footnotesize
\setlength{\tabcolsep}{3pt}
\renewcommand\arraystretch{1.3}

    \caption{Imbalance node classification results of PNS compared with other baselines. We set the imbalance ratio in an extreme setting ($\rho = 80$) with standard errors five times.}
\label{table: Amazon}

\begin{tabular}{ll|lll|lll|lll}
\hline
\multirow{2}{*}{\textbf{}} &
  \textbf{Dataset} &
  \multicolumn{3}{c|}{\textbf{Photo-ST}} &
  \multicolumn{3}{c|}{\textbf{Computer-ST}} &
  \multicolumn{3}{c}{\textbf{CS-ST}} \\ \cline{2-11} 
 &
  $\rho$=80 &
  \multicolumn{1}{c}{Acc.} &
  \multicolumn{1}{c}{bAcc.} &
  \multicolumn{1}{c|}{F1} &
  \multicolumn{1}{c}{Acc.} &
  \multicolumn{1}{c}{bAcc.} &
  \multicolumn{1}{c|}{F1} &
  \multicolumn{1}{c}{Acc.} &
  \multicolumn{1}{c}{bAcc.} &
  \multicolumn{1}{c}{F1} \\ \hline
\multicolumn{1}{c}{\multirow{9}{*}{\rotatebox{90}{GCN}}} &
  Vanilla &
  38.85\scriptsize$\pm$0.09 &
  46.82\scriptsize$\pm$0.09  &
  31.08\scriptsize$\pm$0.46 &
  60.25\scriptsize$\pm$0.45 &
  48.73\scriptsize$\pm$1.40 &
  33.19\scriptsize$\pm$0.18 &
  36.92\scriptsize$\pm$0.68 &
  52.03\scriptsize$\pm$0.59 &
  26.21\scriptsize$\pm$1.24 \\ \cline{2-11} 
\multicolumn{1}{c}{} &
  Reweight &
  44.41\scriptsize$\pm$4.17 &
  50.12\scriptsize$\pm$2.24 &
  32.77\scriptsize$\pm$2.86 &
  61.87\scriptsize$\pm$0.09 &
  49.62\scriptsize$\pm$0.54 &
  33.18\scriptsize$\pm$0.43 &
  39.32\scriptsize$\pm$0.84 &
  51.35\scriptsize$\pm$1.26 &
  25.47\scriptsize$\pm$2.05 \\
\multicolumn{1}{c}{} &
  Upsampling &
  42.98\scriptsize$\pm$3.84 &
  48.96\scriptsize$\pm$1.98 &
  32.19\scriptsize$\pm$2.15 &
  61.77\scriptsize$\pm$0.12  &
  50.06\scriptsize$\pm$0.56 &
  33.29\scriptsize$\pm$0.28 &
  39.66\scriptsize$\pm$2.25 &
  52.39\scriptsize$\pm$1.67 &
  27.80\scriptsize$\pm$2.73 
\\
  \multicolumn{1}{c}{} &
  M-ILBO &
14.99\scriptsize$\pm$8.06 &
24.91\scriptsize$\pm$7.60 &
9.42\scriptsize$\pm$5.49 &
10.94\scriptsize$\pm$10.41 &
13.89\scriptsize$\pm$3.89 &
4.24\scriptsize$\pm$4.13 &
29.71\scriptsize$\pm$0.09 &
43.58\scriptsize$\pm$0.08 &
20.73\scriptsize$\pm$0.10 \\

\multicolumn{1}{c}{} &
  GraphSMOTE &
  45.91\scriptsize$\pm$3.79 &
  50.40\scriptsize$\pm$1.96 &
  32.09\scriptsize$\pm$2.41 &
  61.83\scriptsize$\pm$0.09 &
  50.47\scriptsize$\pm$0.38 &
  33.78\scriptsize$\pm$0.40 &
  39.86\scriptsize$\pm$1.37 &
  52.23\scriptsize$\pm$0.85 &
  26.56\scriptsize$\pm$1.28 \\
\multicolumn{1}{c}{} &
  GraphENS &
  81.17\scriptsize$\pm$0.38 &
  83.40\scriptsize$\pm$0.42 &
  78.64\scriptsize$\pm$0.90 &
  75.36\scriptsize$\pm$0.27 &
  82.72\scriptsize$\pm$0.16 &
  69.72\scriptsize$\pm$0.35 &
  85.05\scriptsize$\pm$0.54 &
  85.45\scriptsize$\pm$0.30 &
  74.68\scriptsize$\pm$0.34 \\
\multicolumn{1}{c}{} &
  PC Softmax &
  37.69\scriptsize$\pm$0.11 &
  45.43\scriptsize$\pm$0.06 &
  33.88\scriptsize$\pm$0.08 &
  43.28\scriptsize$\pm$0.34 &
  34.38\scriptsize$\pm$0.28 &
  29.66\scriptsize$\pm$0.40 &
  38.47\scriptsize$\pm$0.36 &
  49.97\scriptsize$\pm$0.20 &
  33.98\scriptsize$\pm$0.46 \\
\multicolumn{1}{c}{} &
  BAT &
  38.47\scriptsize$\pm$0.19 &
  46.41\scriptsize$\pm$0.11 &
  30.34\scriptsize$\pm$0.50 &
  61.15\scriptsize$\pm$0.09 &
  45.64\scriptsize$\pm$0.06 &
  26.82\scriptsize$\pm$0.05 &
  32.53\scriptsize$\pm$0.07 &
 47.84\scriptsize$\pm$0.24 &
  18.74\scriptsize$\pm$0.39 \\ 
 &

  GNNCL &
38.68\scriptsize$\pm$0.31 &
46.63\scriptsize$\pm$0.32 &
29.07\scriptsize$\pm$1.10 &

49.74\scriptsize$\pm$7.97 &
37.11\scriptsize$\pm$5.45 & 
22.60\scriptsize$\pm$4.13 &

31.83\scriptsize$\pm$0.21 &
46.73\scriptsize$\pm$0.24 &
19.24\scriptsize$\pm$0.99 \\

 &
 
  GraphSHA &
  \textbf{83.22\scriptsize$\pm$0.27} &
  83.05\scriptsize$\pm$0.09 &
  78.52\scriptsize$\pm$0.26 &
  75.66\scriptsize$\pm$0.18 &
  \textbf{83.04\scriptsize$\pm$0.15} &
  \textbf{71.94\scriptsize$\pm$0.18} &
  87.14\scriptsize$\pm$0.10 &
  \textbf{87.63\scriptsize$\pm$0.15} &
  76.34\scriptsize$\pm$0.17  \\ \cline{2-11}
\multicolumn{1}{c}{} &

  \textbf{PNS + BAT} &
  38.61\scriptsize$\pm$0.19&
  46.44\scriptsize$\pm$0.06&
  29.65\scriptsize$\pm$0.42 &
  61.25\scriptsize$\pm$0.15 &
  45.60\scriptsize$\pm$0.10 &
  26.94\scriptsize$\pm$0.17&
  32.79\scriptsize$\pm$0.21 &
  47.95\scriptsize$\pm$0.26 &
  19.03\scriptsize$\pm$0.46\\ 
\multicolumn{1}{c}{} &

  \textbf{PNS + ENS} &
  82.01\scriptsize$\pm$0.67 &
  \textbf{83.34\scriptsize$\pm$0.68} &
  \textbf{80.13\scriptsize$\pm$0.76} &
  \textbf{76.08\scriptsize$\pm$0.56} &
  82.65\scriptsize$\pm$0.31 &
  70.87\scriptsize$\pm$0.26 &
  \textbf{87.49\scriptsize$\pm$0.52} &
  87.09\scriptsize$\pm$0.25 &
  \textbf{76.68\scriptsize$\pm$0.12} \\ \hline
\multirow{9}{*}{\rotatebox{90}{GAT}} &
  Vanilla &
  39.55\scriptsize$\pm$0.04 &
  47.54\scriptsize$\pm$0.04 &
  25.33\scriptsize$\pm$0.09 &
  62.66\scriptsize$\pm$0.31 &
  48.56\scriptsize$\pm$1.88 &
  32.62\scriptsize$\pm$3.11 &
  43.66\scriptsize$\pm$2.82 &
  58.14\scriptsize$\pm$2.47 &
  34.99\scriptsize$\pm$4.01 \\ \cline{2-11} 
 &
  Reweight &
  61.78\scriptsize$\pm$0.5 &
  60.73\scriptsize$\pm$0.09&
  47.89\scriptsize$\pm$0.26 &
  62.92\scriptsize$\pm$0.05 &
  46.80\scriptsize$\pm$0.06 &
  29.72\scriptsize$\pm$0.11 &
  43.62\scriptsize$\pm$1.52 &
  54.44\scriptsize$\pm$1.28 &
  30.40\scriptsize$\pm$2.01 \\
 &
  Upsampling &
  62.32\scriptsize$\pm$0.24 &
  61.35\scriptsize$\pm$0.31 &
  49.40\scriptsize$\pm$0.78 &
  62.85\scriptsize$\pm$0.04 &
  46.76\scriptsize$\pm$0.03 &
  29.62\scriptsize$\pm$0.07 &
  40.16\scriptsize$\pm$2.06 &
  53.12\scriptsize$\pm$1.45&
  27.11\scriptsize$\pm$2.00 \\
 &
 M-ILBO &
34.92\scriptsize$\pm$0.30 &
42.96\scriptsize$\pm$0.37 &
24.02\scriptsize$\pm$0.41 &
54.36\scriptsize$\pm$0.40 &
36.36\scriptsize$\pm$0.34 &
22.81\scriptsize$\pm$0.35 &
28.86\scriptsize$\pm$0.11 &
42.37\scriptsize$\pm$0.08 &
19.86\scriptsize$\pm$0.23 \\

 &
  GraphSMOTE &
  62.41\scriptsize$\pm$0.24 &
  61.53\scriptsize$\pm$0.65&
  49.95\scriptsize$\pm$1.58&
  62.84\scriptsize$\pm$0.02 &
  46.87\scriptsize$\pm$0.02 &
  29.64\scriptsize$\pm$0.10 &
  41.37\scriptsize$\pm$2.62 &
  54.89\scriptsize$\pm$1.57 &
  30.13\scriptsize$\pm$2.59 \\
 &
  GraphENS &
  83.45\scriptsize$\pm$0.20 &
  \textbf{85.93\scriptsize$\pm$0.38} &
  82.08\scriptsize$\pm$0.29 &
  76.20\scriptsize$\pm$0.50 &
  \textbf{83.63\scriptsize$\pm$0.34} &
  71.26\scriptsize$\pm$0.65 &
  87.54\scriptsize$\pm$0.31 &
  87.81\scriptsize$\pm$0.28 &
  76.17\scriptsize$\pm$1.41 \\
 &
  PC Softmax &
  49.55\scriptsize$\pm$4.67 &
  58.31\scriptsize$\pm$2.01 &
  48.45\scriptsize$\pm$3.43 &
  33.34\scriptsize$\pm$4.01 &
  49.09\scriptsize$\pm$2.24 &
  34.91\scriptsize$\pm$1.32 &
  73.07\scriptsize$\pm$1.00 &
  71.83\scriptsize$\pm$1.01 &
  59.55\scriptsize$\pm$1.42 \\
 &
   BAT &
  39.03\scriptsize$\pm$0.11 &
  47.08\scriptsize$\pm$0.09 &
  30.78\scriptsize$\pm$0.80 &
  62.92\scriptsize$\pm$0.25 &
  45.98\scriptsize$\pm$0.10 &
  32.07\scriptsize$\pm$0.59 &
  32.30\scriptsize$\pm$0.07 &
  47.25\scriptsize$\pm$0.13 &
  20.93\scriptsize$\pm$1.10 \\ 
 &

  GNNCL &
39.87\scriptsize$\pm$0.71 &
47.90\scriptsize$\pm$0.77 &
30.73\scriptsize$\pm$0.70 &
60.95\scriptsize$\pm$1.85 & 
42.37\scriptsize$\pm$4.25 & 
28.53\scriptsize$\pm$1.96 &
32.79\scriptsize$\pm$0.36 &
47.74\scriptsize$\pm$0.75 &
18.66\scriptsize$\pm$1.02 \\

 &
 
  GraphSHA &
  71.22\scriptsize$\pm$2.89 &
  75.60\scriptsize$\pm$2.68 &
  68.23\scriptsize$\pm$2.64 &
  70.78\scriptsize$\pm$1.59 &
  76.70\scriptsize$\pm$0.87 &
  62.74\scriptsize$\pm$1.85 &
  82.25\scriptsize$\pm$1.12 &
  82.94\scriptsize$\pm$0.58 &
  65.20\scriptsize$\pm$0.92 \\ \cline{2-11}

 &

   \textbf{PNS + BAT} &
  39.16\scriptsize$\pm$0.17 &
  47.16\scriptsize$\pm$0.07&
  31.60\scriptsize$\pm$0.26 &
  63.26\scriptsize$\pm$0.29&
  46.22\scriptsize$\pm$0.15&
  33.14\scriptsize$\pm$0.30&
  32.54\scriptsize$\pm$0.09&
  47.42\scriptsize$\pm$0.28&
  18.96\scriptsize$\pm$0.66\\ 
\multicolumn{1}{c}{} &

  \textbf{PNS + ENS} &
  \textbf{83.78\scriptsize$\pm$0.35} &
  85.19\scriptsize$\pm$0.24 &
  \textbf{82.47\scriptsize$\pm$0.27} &
  \textbf{76.75\scriptsize$\pm$0.25} &
  83.38\scriptsize$\pm$0.12 &
  \textbf{72.19\scriptsize$\pm$0.31} &
  \textbf{89.31\scriptsize$\pm$0.46} &
  \textbf{88.74\scriptsize$\pm$0.19} &
  \textbf{78.97\scriptsize$\pm$0.35} \\ \hline
\multirow{8}{*}{\rotatebox{90}{SAGE}} &
  Vanilla &
 43.37\scriptsize$\pm$1.51  &
 49.62\scriptsize$\pm$0.78 &
 29.58\scriptsize$\pm$1.35  &
  63.46\scriptsize$\pm$0.19 &
  46.89\scriptsize$\pm$0.11 &
  29.91\scriptsize$\pm$0.17 &
  43.49\scriptsize$\pm$1.04 &
  56.44\scriptsize$\pm$0.66 &
  32.04\scriptsize$\pm$1.15 \\ \cline{2-11} 
 &
  Reweight &
  54.31\scriptsize$\pm$2.35 &
  56.52\scriptsize$\pm$0.95 &
  41.88\scriptsize$\pm$1.75 &
  61.47\scriptsize$\pm$0.62 &
  49.12\scriptsize$\pm$1.32 &
  34.27\scriptsize$\pm$0.20 &
  53.18\scriptsize$\pm$1.39&
  61.52\scriptsize$\pm$0.78 &
  40.91\scriptsize$\pm$1.43 \\
 &
  Upsampling &
  56.29\scriptsize$\pm$1.16 &
  57.67\scriptsize$\pm$1.20 &
  43.97\scriptsize$\pm$1.70 &
  62.80\scriptsize$\pm$0.03 &
  52.40\scriptsize$\pm$1.90 &
  37.24\scriptsize$\pm$2.38&
  51.99\scriptsize$\pm$1.99 &
  62.63\scriptsize$\pm$0.64 &
  41.62\scriptsize$\pm$1.13 \\
 &
 M-ILBO &
1.83\scriptsize$\pm$0.00 &
12.50\scriptsize$\pm$0.00 &
0.45\scriptsize$\pm$0.00 &
0.53\scriptsize$\pm$0.00 &
10.00\scriptsize$\pm$0.00 &
0.11\scriptsize$\pm$0.00 &
2.34\scriptsize$\pm$0.00 &
7.14\scriptsize$\pm$0.00 &
0.33\scriptsize$\pm$0.00 \\

 &
  GraphSMOTE &
  56.68\scriptsize$\pm$1.27 &
  56.24\scriptsize$\pm$0.54&
  41.52\scriptsize$\pm$1.04&
  61.35\scriptsize$\pm$0.45 &
  48.73\scriptsize$\pm$1.40 &
  33.19\scriptsize$\pm$1.80 &
  53.08\scriptsize$\pm$1.72&
  63.64\scriptsize$\pm$0.88 &
   40.93\scriptsize$\pm$1.19\\
 &
  GraphENS &
  81.87\scriptsize$\pm$0.38 &
  83.40\scriptsize$\pm$0.42 &
  78.64\scriptsize$\pm$0.90 &
  72.51\scriptsize$\pm$0.35 &
  79.16\scriptsize$\pm$0.71 &
  65.44\scriptsize$\pm$0.72 &
  85.61\scriptsize$\pm$0.55 &
  86.53\scriptsize$\pm$0.24 &
  75.05\scriptsize$\pm$0.74 \\
 &
  PC Softmax &
  67.23\scriptsize$\pm$0.79 &
  74.96\scriptsize$\pm$0.38 &
  66.73\scriptsize$\pm$0.62 &
  73.30\scriptsize$\pm$0.08 &
  60.51\scriptsize$\pm$0.29 &
  53.56\scriptsize$\pm$0.30 &
  \textbf{88.58\scriptsize$\pm$0.18} &
  \textbf{88.31\scriptsize$\pm$0.17} &
  \textbf{77.66\scriptsize$\pm$1.21} \\
 &
  BAT &
  45.99\scriptsize$\pm$2.04 &
  53.04\scriptsize$\pm$1.22 &
  35.04\scriptsize$\pm$2.15 &
 63.01\scriptsize$\pm$0.60 &
  46.83\scriptsize$\pm$0.29 &
  30.43\scriptsize$\pm$0.22 &
  63.01\scriptsize$\pm$0.60 &
  46.83\scriptsize$\pm$0.29 &
  30.43\scriptsize$\pm$0.22 \\ 
 &

  GNNCL &
39.44\scriptsize$\pm$0.09 &
47.10\scriptsize$\pm$0.10 & 
28.56\scriptsize$\pm$0.78 &
60.10\scriptsize$\pm$1.21 & 
44.82\scriptsize$\pm$0.91 & 
26.83\scriptsize$\pm$0.54 &
39.48\scriptsize$\pm$1.23 & 
52.97\scriptsize$\pm$0.52 &
28.06\scriptsize$\pm$1.06 \\

 &
 
  GraphSHA &
  \textbf{83.93\scriptsize$\pm$0.13} &
  \textbf{84.93\scriptsize$\pm$0.17} &
  \textbf{79.88\scriptsize$\pm$0.33} &
  \textbf{76.86\scriptsize$\pm$0.17} &
  \textbf{84.98\scriptsize$\pm$0.10} &
  \textbf{72.53\scriptsize$\pm$0.34} &
  87.27\scriptsize$\pm$0.11 &
  88.17\scriptsize$\pm$0.14 &
  73.54\scriptsize$\pm$1.23 \\ \cline{2-11}
 &
   \textbf{PNS + BAT} &
  51.12\scriptsize$\pm$2.82 &
  57.44\scriptsize$\pm$2.18&
  41.93\scriptsize$\pm$3.48 &
  63.72\scriptsize$\pm$0.26 &
  47.14\scriptsize$\pm$0.13 &
 30.46\scriptsize$\pm$0.25&
  53.75\scriptsize$\pm$1.26 &
  64.32\scriptsize$\pm$0.69 &
  42.54\scriptsize$\pm$0.82\\ 
\multicolumn{1}{c}{} &

  \textbf{PNS + ENS} &
  82.06\scriptsize$\pm$1.08&
  83.36\scriptsize$\pm$0.71&
  79.80\scriptsize$\pm$1.02 &
  71.80\scriptsize$\pm$0.56
  &
  76.94\scriptsize$\pm$0.69
  &
  64.17\scriptsize$\pm$1.01
  &
  87.52\scriptsize$\pm$0.37 &
  87.25\scriptsize$\pm$0.14 &
  75.96\scriptsize$\pm$1.07 \\ \hline
\end{tabular}

\end{table*}

In this section, we introduce the experimental setup, analyze the experiments conducted on all datasets, and examine the influence of hyperparameters.

\subsection{Experimental Settings}

\begin{table}[h]
    \footnotesize
    \setlength{\tabcolsep}{2pt}
    \renewcommand\arraystretch{1.5}
    \caption{Parameter settings.}
    \label{table: parameter setting}

\begin{tabular}{ll|llllll}
\hline
     &         & \textbf{Cora} & \textbf{Citeseer} & \textbf{PubMed} & \textbf{Photo} & \textbf{Computer} & \textbf{CS}  \\ \hline
\multirow{2}{*}{\rotatebox{90}{GCN}}  & PNS+ENS & 0.3  & 0.5      & 0.7    & 0.7   & 0.7      & 0.7 \\
     & PNS+BAT & 0.6  & 0.6      & 0.7    & 0.5   & 0.7      & 0.7 \\ \hline
\multirow{2}{*}{\rotatebox{90}{GAT}}  & PNS+ENS & 0.3  & 0.4      & 0.7    & 0.7   & 0.7      & 0.7 \\
     & PNS+BAT & 0.5  & 0.9      & 0.7    & 0.7   & 0.3      & 0.6 \\ \hline
\multirow{2}{*}{\rotatebox{90}{SAGE}} & PNS+ENS & 0.3  & 0.3      & 0.7    & 0.7   & 0.1      & 0.7 \\
     & PNS+BAT & 0.8  & 0.01     & 0.7    & 0.7   & 0.7      & 0.7 \\ \hline
\end{tabular}

\end{table}

The confusion rate setting in different datasets can be referred to Table \ref{table: parameter setting}. All GNN backbone settings are the same, with layer = 2 and feature dimension = 256. It should be noted that the heads in GAT are 8.

\textbf{Datasets.} We validate PNS on six benchmark datasets, which are citation networks \citep{sen2008collective} and Amazon dataset \citep{shchur2018pitfalls}. For citation network datasets, including Cora,  a citation network where each node represents a paper, and edges represent citation relationships between papers; CiteSeer, another citation network, similar to Cora but with a different scale and characteristics; and PubMed, a larger citation network of papers related to biomedical research.
For Amazon datasets, including Photo, a product co-purchasing network derived from Amazon; Computer, Another co-purchasing network from Amazon, similar to the Photo dataset; and CS, a collaboration network derived from the Microsoft academic graph. We adopt the LT setting for citation networks \citep{park2021graphens} and step-setting for Amazon datasets \citep {li2023graphsha}. The data details are shown in Table \ref{table: dataset deatils}.

\textbf{Baselines.} We compare PNS with six different baselines and test it using three different GNN backbones, which are GCN \citep{kipf2016semi}, GAT \citep{velivckovic2017graph}, and GraphSAGE \citep{hamilton2017inductive}. We compare (1) Reweight, which weights classes proportionally by quantity; (2) Upsampling, which duplicates minor nodes directly; (3) GraphSMOTE \citep{zhao2021graphsmote}, which synthesizes the minors by two minor nodes in the same class; (4) GraphENS \citep{park2021graphens}, which synthesizes new minor nodes by combining ego-networks of minor nodes with random nodes; (5) PC softmax \citep{hong2021disentangling}, which modifies the loss to solve the imbalance problem; (6) GraphSHA \citep{li2023graphsha}, which synthesizes the minor nodes using hard samples. (7) M-ILBO \citep{ma2023entropy} utilizes two random augmented views by contrast learning with a class-balance setting in unsupervised learning. The cross-view guarantees the learned representations are consistent across views from the perspective of the entire graph.  (8) BAT \citep{liuclass} addresses the topological challenges inherent in class-imbalanced graph learning by leveraging dynamic topological augmentation. Noticed that we use BAT1 in our experiment. To ensure a fair comparison, we employ a standardized data input format across all baselines. (9) GNN-CL \citep{li2024graph} incorporates curriculum learning by adaptively generating interpolated nodes and edges, while jointly optimizing graph classification loss and metric learning. For all the baselines, besides GraphSHA, conducting all the experiments in AutoDL with NVIDIA 3090 24 GB GPU, other experiments are in the NVIDIA GeForce RTX 4060 Ti 16 GB GPU. We set the imbalance ratio $\rho$ = 100 when conducting experiments in Cora, PubMed, and Citeseer, and $\rho$ = 80 when conducting experiments in Amazon-Photo, Amazon-computer, and CS. 

\textbf{Evaluation metric.} We first introduce the evaluation metrics used for different datasets. We employ three evaluation metrics to assess model performance: (1) Accuracy (Acc.) represents the proportion of correctly predicted samples to the total sample; (2) Balanced Accuracy (bAcc.) is a refined metric that addresses class imbalance by calculating the recall rate for each category and then taking the average value; and (3) F1 Score (F1) is the harmonic mean of Precision and Recall, which evaluates the model's ability to balance true positives and false positives, providing a comprehensive measure of its performance on both positive and negative samples. We evaluated the performance of various baselines across all datasets using the three metrics mentioned above: Accuracy, Balanced Accuracy, and F1 Score.

\subsection{Main Results}

We follow the dataset setting method in both GraphENS and GraphSHA. We utilize the LT class imbalance setting \citep{park2021graphens} for datasets Cora, Citseer, and PubMed. In addition, we utilize the step class imbalance setting \citep{chen2021topology, zhang2022m} for Amazon photo, computer, and CS in our whole experiment. In the LT class imbalance setting datasets, we set $\rho$ = 100. In step class imbalance setting datasets, we set $\rho$ = 80. The value of $r$ is calibrated specifically for each dataset and backbone.

In Table \ref{table: citation networks}, we find that PNS demonstrates improvement over other baselines across different GNN backbones. PNS achieves SOTA performance, demonstrating its effectiveness. In other words, these datasets may contain more nodes with anomalous connectivity problems. PNS also achieves better performance than the original GraphENS and BAT. Additionally, we observe that some methods perform even worse than Vanilla. For example, on Cora-LT: With GCN as the backbone, Reweight, Upsampling, and GraphSMOTE achieve performance drops of 1.22$\% \downarrow$, 1.94$\% \downarrow$, 2.12$\% \downarrow$. With GAT, Upsampling results in a 0.6$\% \downarrow$ performance drop, while Reweight and GraphSMOTE perform slightly better than the Vanilla approach. With SAGE, Reweight, Upsampling, and GraphSMOTE all slightly outperform the Vanilla approach. Among these methods, PNS emerges as the top-performing approach, outdoing Vanilla across different GNN backbones.

In Table \ref{table: Amazon}, we find that PNS also shows excellent performance in Amazon-related datasets, while BAT experiences a performance degradation. PNS achieves outstanding performance when paired with GCN and GAT as backbone architectures. When using GCN as the backbone, PNS achieves top performance on Computer-ST and CS-ST in accuracy. Notably, while GraphSHA yields the highest Acc. on Photo-ST, PNS surpasses it in both balanced accuracy and F1 score. PNS achieves the best performance with GAT as the backbone across three datasets. By utilizing GraphSAGE as its backbone, although PNS exhibits commendable performance, it does not occupy the top spot, instead ranking second overall. In particular, PNS experiences a worse performance compared with the original GraphENS when using SAGE as the GNN backbone. GraphSHA achieves the best performance on Photo-ST and Computer-ST, outperforming PNS by 1.87\% and 5.06\% in terms of accuracy, respectively. Notably, PC Softmax achieves the best performance on CS-ST, with PNS trailing behind by 1.06\% in terms of accuracy. The results likely suggest that there are either few nodes with anomalous connectivity problems, or that the connection patterns of nodes in Amazon differ from those in other datasets. In almost all datasets, PNS has better performance than GraphENS and BAT. The experiment shows that PNS is an effective method. Noticed that M-ILBO fails in an imbalance situation, especially with SAGE as a backbone. The random sampling strategy of M-ILBO can disrupt the structure of the graph in imbalanced situations, since both the quantity and topological structure of the minority class are extremely rare. In contrast, PNS preserves the overall graph structure under imbalance.

\begin{figure}
    \centering
    \includegraphics[scale=0.41]{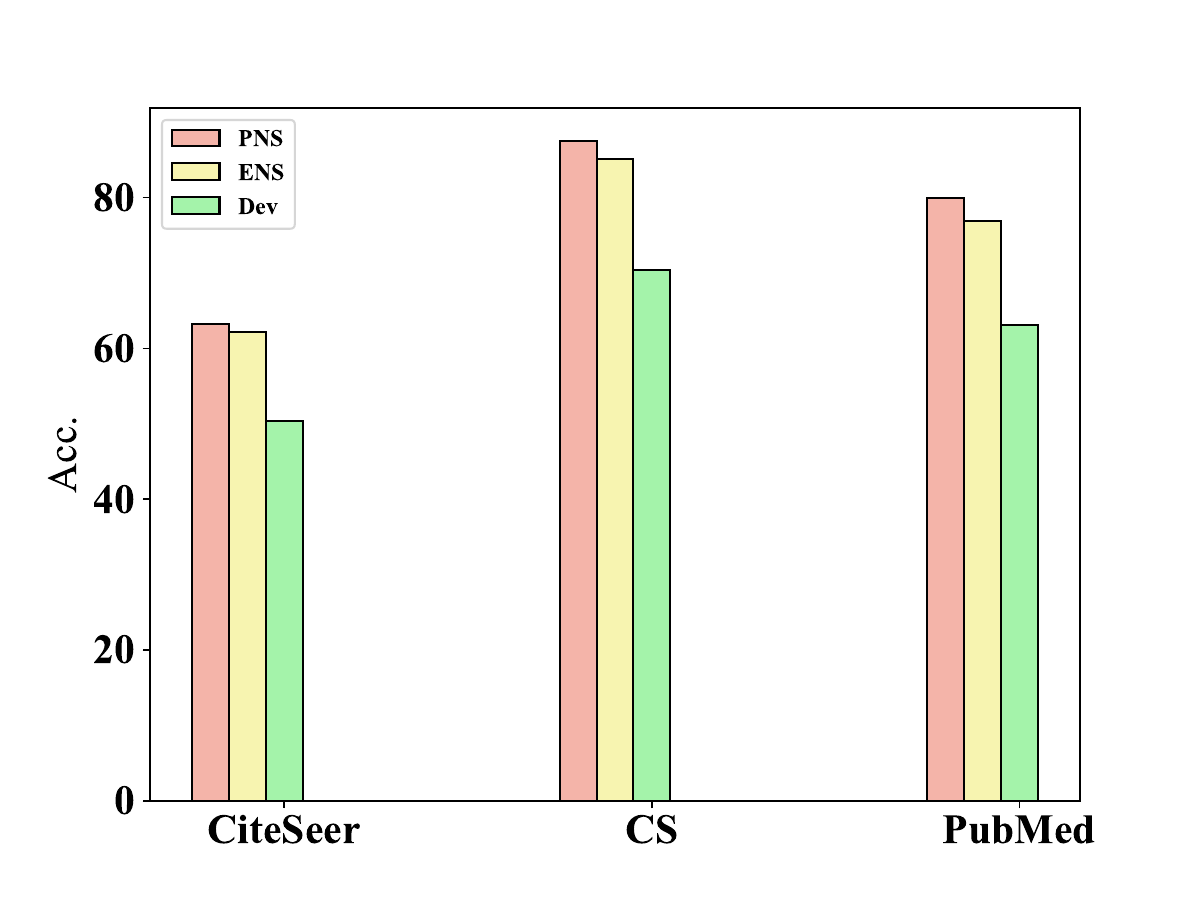}
    \caption{Ablation study on three different datasets.}
    \label{fig: ablation}
\end{figure}

\begin{table}[!t]
\centering
\begin{center}
\caption{Node classification results ($\pm$std) on large-scale naturally class-imbalanced dataset ogbn-arXiv. OOM indicates Out-Of-Memory on a 16GB GPU.}\label{table:ogbn}
\vspace{-0.2cm}
\scalebox{0.80}{
\begin{tabular}{l|cccc}
\toprule
{\textbf{Method}} & {Val Acc.} & {Test Acc.} & {Test bAcc.} & {Test F1} \\
\midrule
Vanilla (GCN) & 73.02\footnotesize{$\pm$0.14} & 71.81\footnotesize{$\pm$0.26} & 50.96\footnotesize{$\pm$0.21} & 50.42\footnotesize{$\pm$0.18} \\
\midrule
Reweight & 67.49\footnotesize{$\pm$0.32} & 66.07\footnotesize{$\pm$0.55} & 53.34\footnotesize{$\pm$0.30} & 48.07\footnotesize{$\pm$0.77} \\
PC Softmax & 72.19\footnotesize{$\pm$0.11} & 71.49\footnotesize{$\pm$0.25} & 48.14\footnotesize{$\pm$0.14} & 50.59 \footnotesize{$\pm$0.13} \\
CB Loss & 65.75\footnotesize{$\pm$0.23} & 64.73\footnotesize{$\pm$0.86} & 52.66\footnotesize{$\pm$0.72} & 47.24\footnotesize{$\pm$1.25} \\
Focal Loss & 67.36\footnotesize{$\pm$0.24} & 65.93\footnotesize{$\pm$0.58} & 53.06\footnotesize{$\pm$0.21} & 48.89\footnotesize{$\pm$0.72} \\
ReNode & 66.44\footnotesize{$\pm$0.51} & 65.91\footnotesize{$\pm$0.20} & 53.39\footnotesize{$\pm$0.40} & 48.18\footnotesize{$\pm$0.52} \\
TAM (ReNode) & 67.91\footnotesize{$\pm$0.27} & 66.63\footnotesize{$\pm$0.66} & 
\textbf{53.40\scriptsize$\pm$0.24} & 48.71\footnotesize{$\pm$0.49} \\

Upsample & 70.53\footnotesize{$\pm$0.08} & 69.55\footnotesize{$\pm$0.37} & 46.82\footnotesize{$\pm$0.07} & 45.49\footnotesize{$\pm$0.20} \\
GraphSmote & OOM & OOM & OOM & OOM \\
GraphENS & OOM & OOM & OOM & OOM \\
\midrule
\textbf{PNS + GCN} & 
\textbf{74.01\scriptsize$\pm$0.01}& 
\textbf{72.31\scriptsize$\pm$0.35}& 
50.36\scriptsize$\pm$1.12& 
\textbf{52.56\scriptsize$\pm$0.53}\\
\bottomrule
\end{tabular}
}
\end{center}
\vspace{-0.5cm}
\end{table}

\subsection{Results on Large-scale Graph}

The class imbalance problem is also a common issue in real-life large-scale graphs. We explore the performance of PNS in a large-scale naturally imbalanced graph to validate the generalization. We conduct our experiment on the OGB-Arxiv dataset \citep{hu2020ogb}, which exhibits a severe class imbalance, with an imbalance ratio of 775 in the training set, 2,282 in the validation set, 2,148 in the test set, and an overall imbalance ratio of 942. We conduct PNS with GCN, and the confusion rate equals 0.3. The other performances are original from GraphSHA \citep{li2023graphsha} and the OGB leaderboard \citep{hu2020ogb}. The result is illustrated in Table \ref{table:ogbn}. OOM denotes that the model is out of memory. We report the validation accuracy, test accuracy, balanced accuracy, and the test F1 in the experiment. It can be observed that, except for test BAcc., PNS outperforms all baseline methods in the experiments. These results demonstrate the effectiveness of PNS on large-scale graphs. Notably, the function of PNS in this experiment is to mitigate performance degradation caused by the abnormal distribution of node neighbors since GCN doesn't incorporate the node synthesis stage.

\subsection{Ablation Study}

We perform an ablation study on three datasets: CiteSeer, CS, and PubMed, using GCN as the GNN backbone. We conducted experiments to evaluate the effectiveness of PNS in addressing the RACP problem, that is, avoiding the selection of nodes with anomalous connectivity (AC) issues for synthesis. Moreover, we explore the influence of the confusion rate $r$ on the model performance on the PubMed dataset. To establish a baseline, we compare methods that only use GraphENS with those that use either pure nodes or confused nodes exclusively in the synthesis process. The results are presented in Fig. \ref{fig: ablation}, where PNS represents the use of PNS, ENS denotes the exclusive use of GraphENS, and Dev indicates the use of only confused nodes for synthesis. As shown in Fig. \ref{fig: ablation}, selecting only nodes with anomalous connectivity (AC) issues would have significantly degraded the model's performance; Using only the original GraphENS yields decent but suboptimal performance; Synthesizing with pure nodes yields the optimal performance. For instance, on Citeseer, Dev suffers a 12.94\% performance drop compared to GraphENS. PNS achieves performance gains of 1.04\% and 12.94\% over GraphENS and Dev, respectively. In summary, PNS effectively addresses the RACP problem by avoiding the selection of nodes with anomalous connectivity (AC) during node synthesis.

\subsection{Influence of Confusion Rate} 

\begin{figure*}[t]
    \centering
    \subfigure[Accuracy]{
        \label{fig: confusion ratio acc}
        \includegraphics[scale = 0.33]{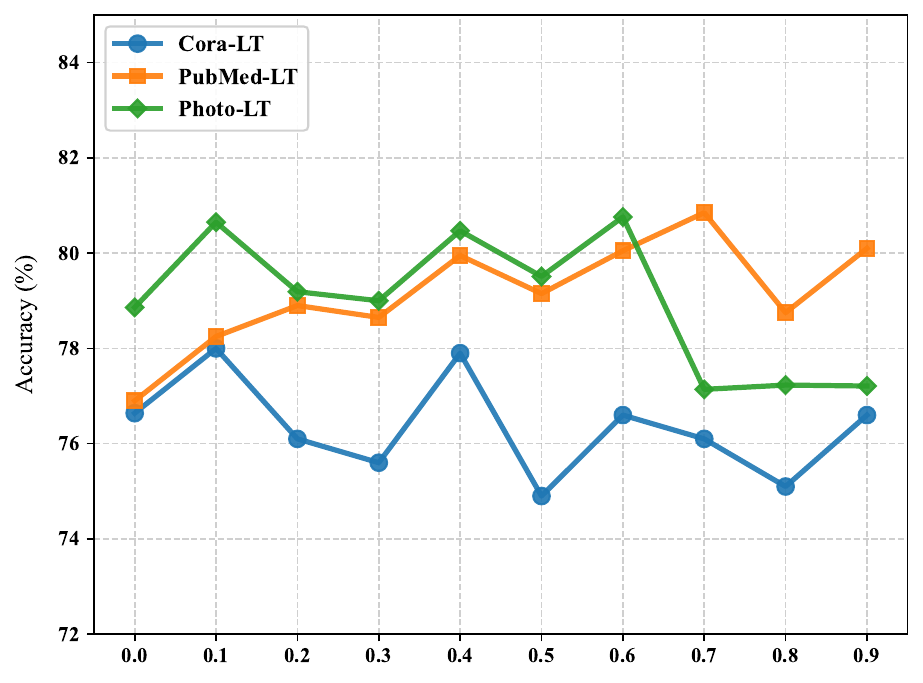}
    }
    \subfigure[Balanced accuracy]{
        \label{ig: confusion ratio bacc}
        \includegraphics[scale = 0.33]{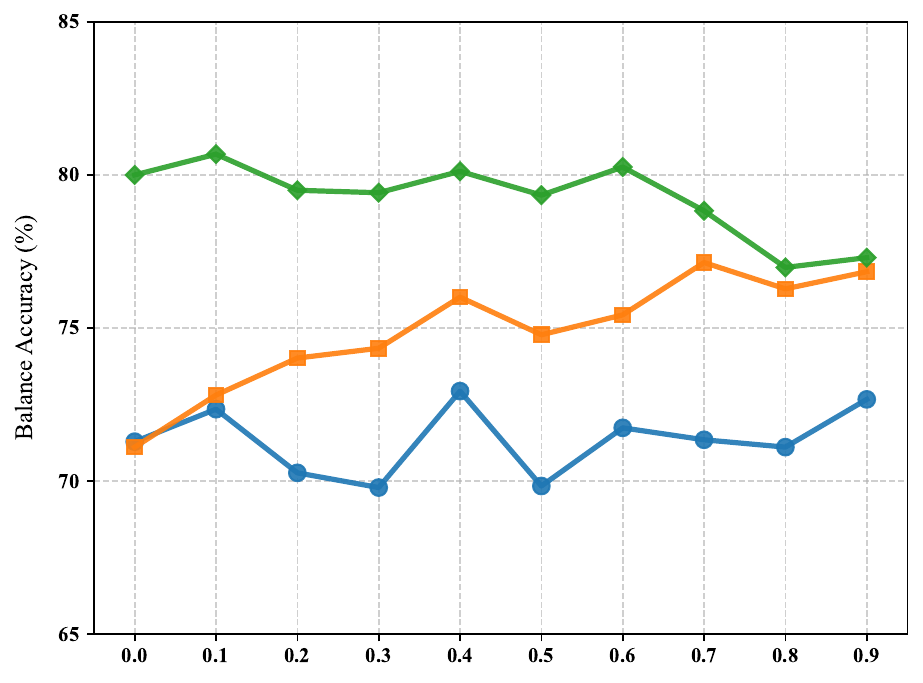}
    }
    \subfigure[F1 score]{
        \label{ig: confusion ratio f1}
        \includegraphics[scale = 0.33]{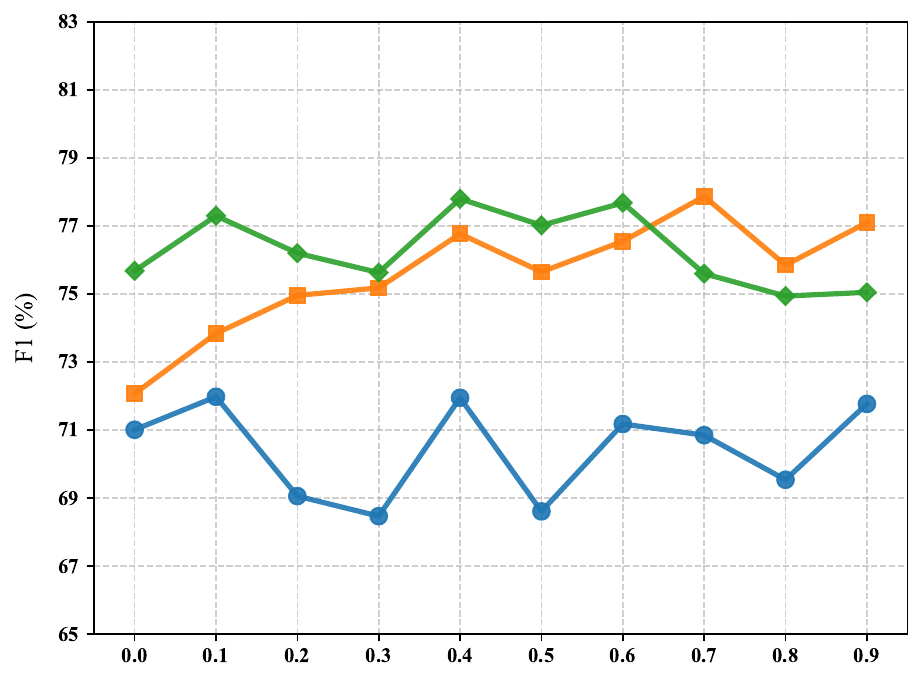}
    }
    \caption{The experiment results of analyzing the influence confusion rate. We conduct the experiment with $\rho$ = 100 with different confusion rates from 0 to 0.9 in a supervised setting, using Cora-LT, PubMed-LT, and Photo-ST with the GCN backbone. We present the tendency of different datasets in accuracy, balanced accuracy, and F1 score as $r$ increases. }
    \label{fig: confusion ratio curve}
\end{figure*}

We study the effect of the confusion rate $r$ in this section. We analyze the performance of PNS using a fixed imbalance ratio $\rho$ = 100 while varying the confusion rate from 0 to 0.9. The results are evaluated on the Cora-LT, PubMed-LT, and Photo-ST datasets using the GCN backbone. Complete experimental results analyzing the influence of the confusion rate are shown in Fig. \ref{fig: confusion ratio curve}. In terms of accuracy, we observe that overall performance initially increases, reaching a peak across all datasets as the confusion rate $r$ increases. Beyond this peak, the performance gradually declines. Overall, the trend shows an initial improvement followed by a subsequent decrease. In terms of BAcc., we observe that for Cora-LT and Photo-ST, BAcc. initially increases and then decreases as the confusion rate $r$ rises. In contrast, for PubMed-LT, BAcc. continues to increase with increasing $r$. In terms of the F1 score, a trend similar to that observed for BAcc. is evident. The reported performance reflects changes in $r$ according to the specific data distribution. For datasets containing a substantial number of confusing nodes, PNS is proven to be highly effective. It is important to note that retaining some confusing nodes remains necessary to prevent overfitting.

\begin{figure*}[b]
    \centering
    \subfigure[Perspective of varying imbalance ratios]{
        \label{fig: homo_imb_surface_imb}
        \includegraphics[scale = 0.26]{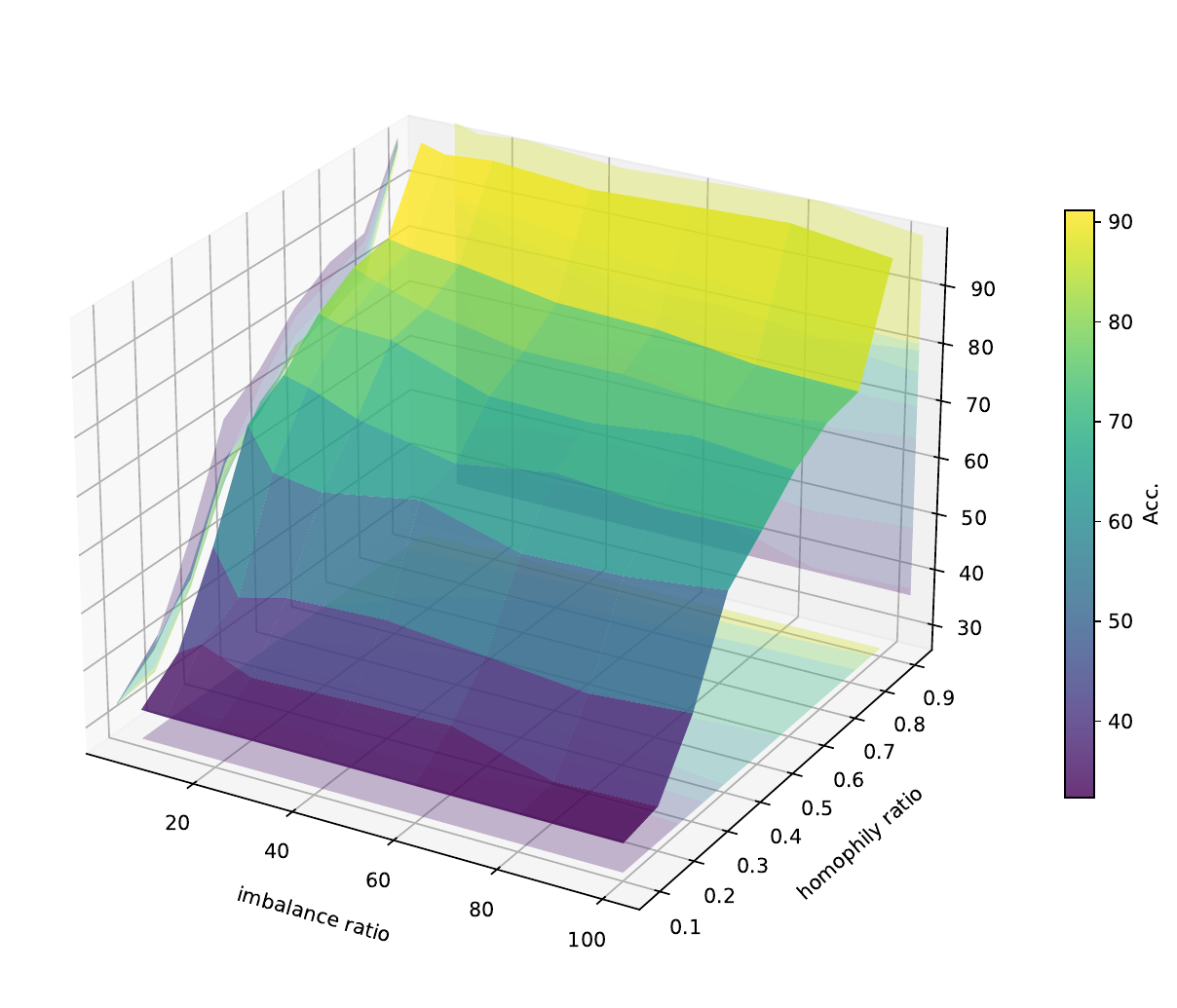}
    }
    \hspace{-6mm}
    \subfigure[Perspective of varying homophily ratio]{
        \label{fig: homo_imb_surface_homo}
        \includegraphics[scale = 0.26]{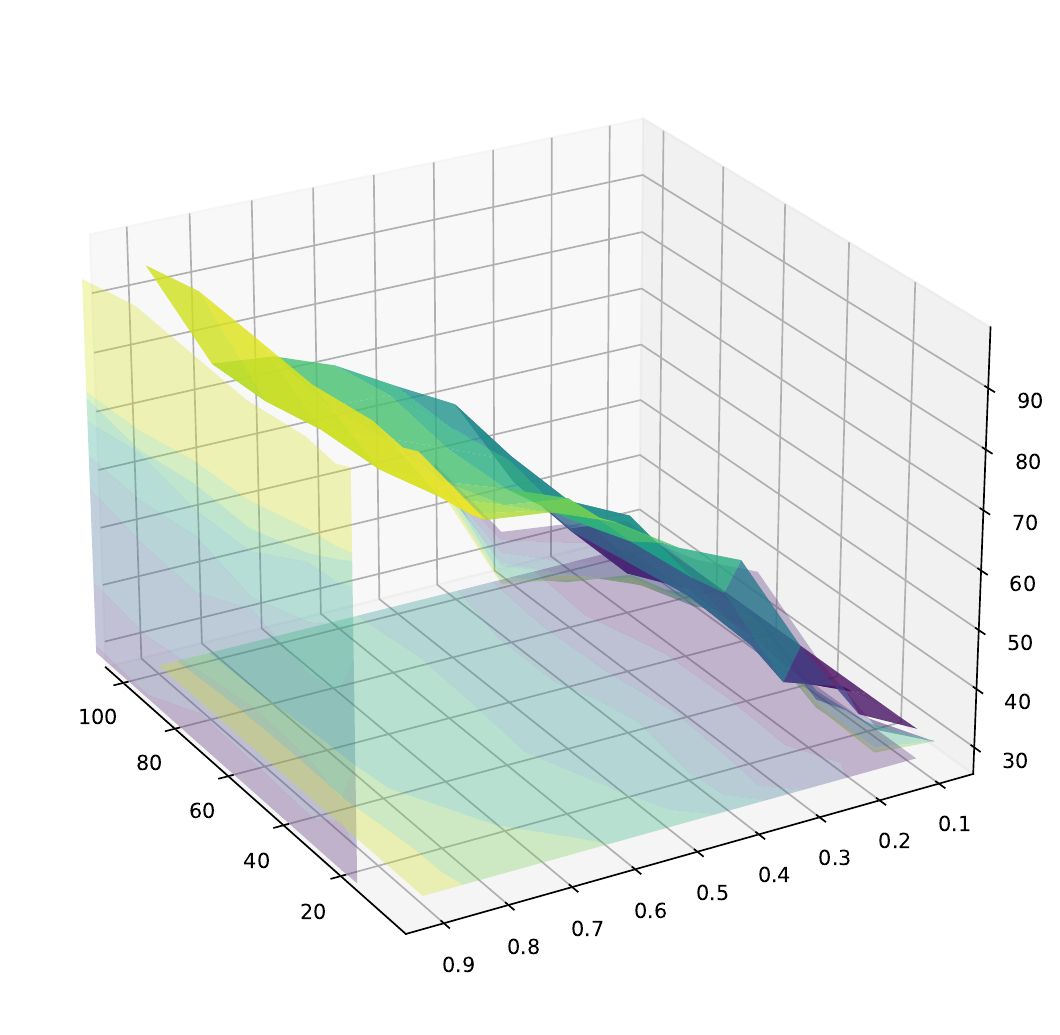}
    }
    \hspace{-6mm}
    \subfigure[Perspective of entire performance]{
        \label{fig: homo_imb_surface_acc}
        \includegraphics[scale = 0.26]{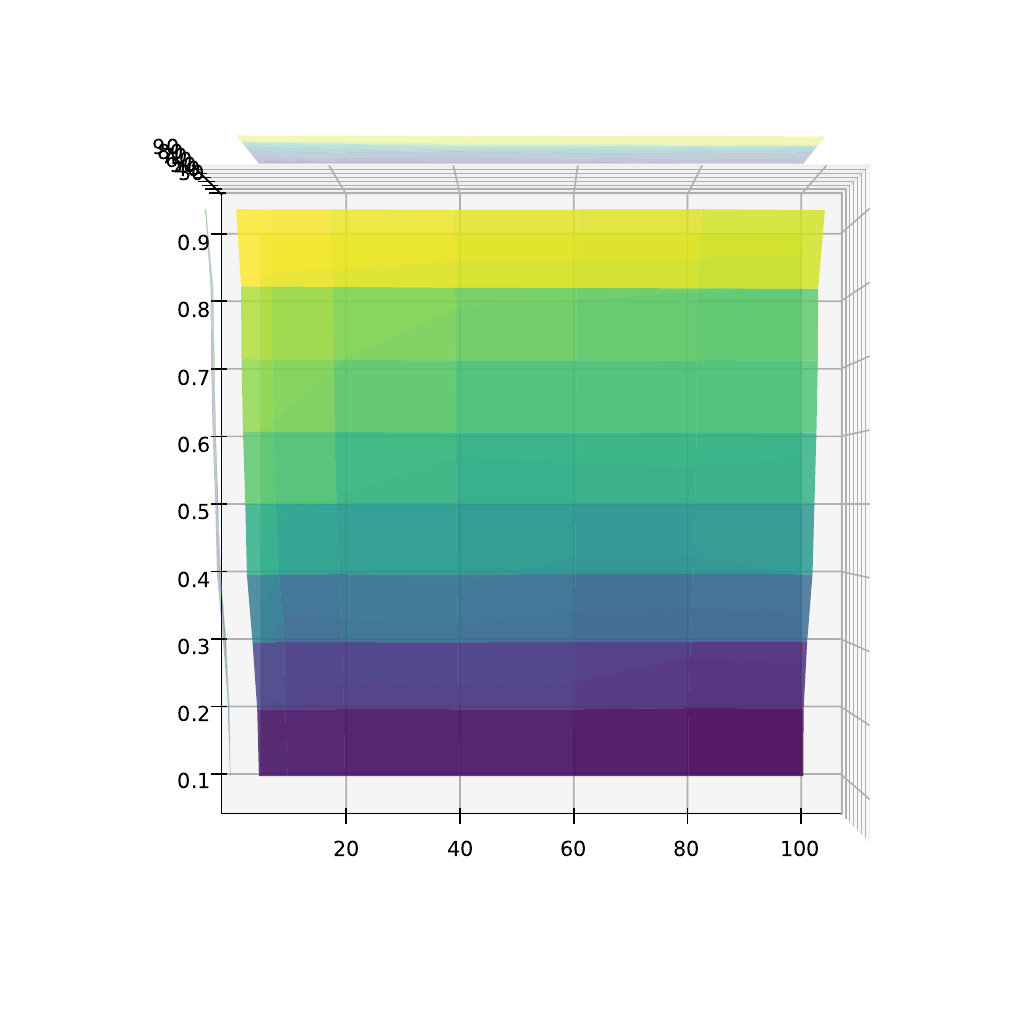}
    }
    \caption{The experiment results of exploring the homophily sensitivity of PNS in varying imbalance ratios. We investigate the combined effects of homophily and class imbalance on the performance of PNS.}
    \label{fig: Homophily sensitivity}
\end{figure*}

\subsection{Homophily Sensitivity}
We investigate the combined effects of homophily and class imbalance on the performance of PNS. The experimental results are presented in Fig. \ref{fig: Homophily sensitivity}. Accuracy is used as the evaluation metric, and the figure is shown from multiple perspectives to provide a clearer understanding of how different factors influence model performance. As illustrated in Fig. \ref{fig: homo_imb_surface_imb}, we can observe that: (1) for a fixed homophily ratio, performance remains relatively consistent across different imbalance ratios. When the imbalance ratio decreases—for instance, at a homophily ratio of 0.2—the color in the figure becomes lighter, indicating a slight improvement in performance; (2) for a fixed imbalance ratio, model performance improves as the homophily ratio increases. As shown in Fig. \ref{fig: homo_imb_surface_homo}, model performance consistently improves with increasing homophily ratio. As shown in Fig. \ref{fig: homo_imb_surface_acc}, in summary, when considering only a single factor, the model achieves higher performance with either a higher homophily ratio or a lower imbalance ratio. When both factors are considered simultaneously, it becomes evident that the homophily ratio has a particularly significant impact on model performance. For instance, at the lowest imbalance ratio ($r$ = 5) and the lowest homophily ratio (0.1), performance in the top-left region is very poor. In contrast, at the highest imbalance ratio ($r$ = 100) and a high homophily ratio (0.9), performance in the top-right region remains excellent.

\begin{figure}
    \centering
    \subfigure[Original dataset]{
        \label{fig: Visualization ori}
        \includegraphics[scale = 0.15]{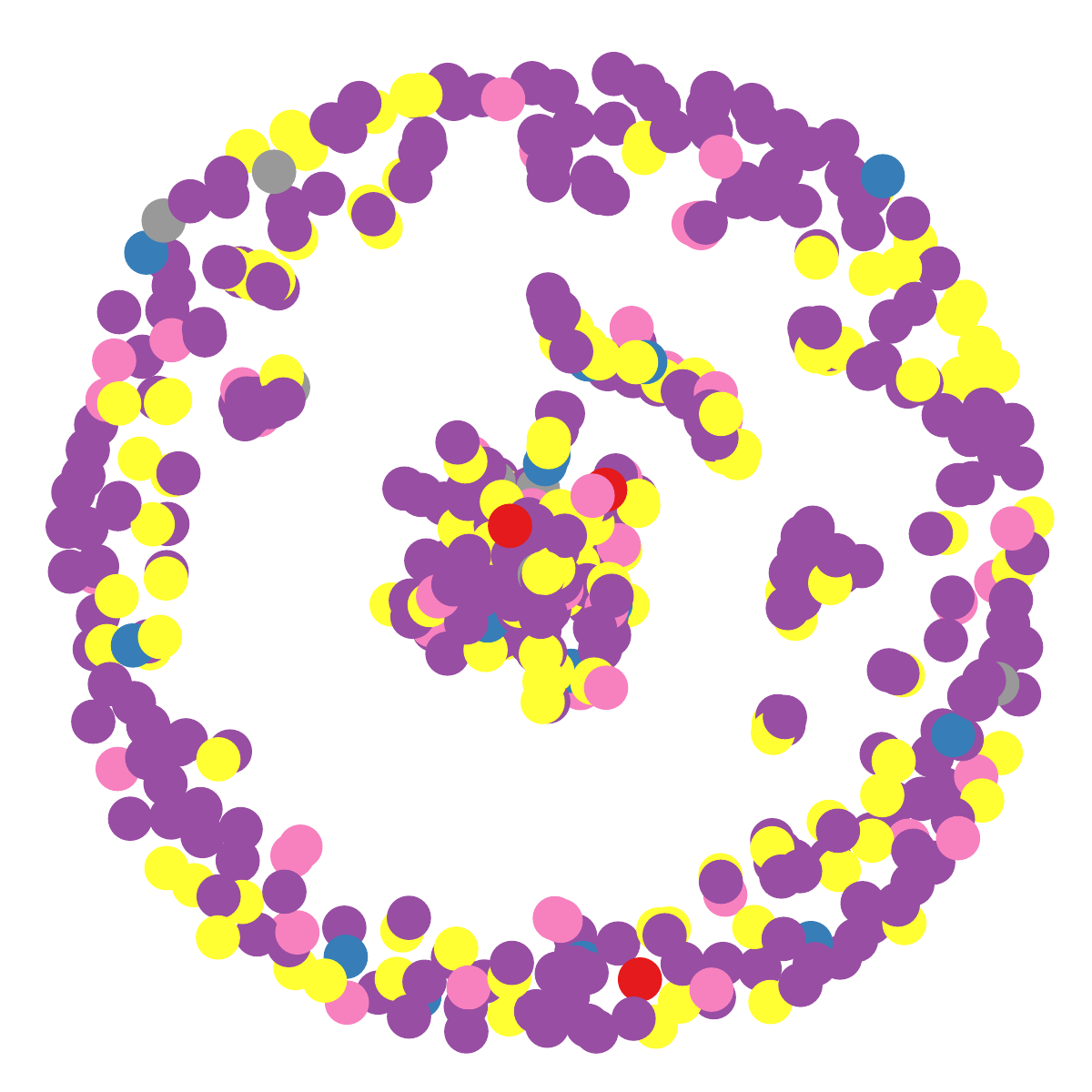}
    }
    \hspace{-4mm}
    \subfigure[Confusion node sets]{
        \label{fig: Visualization dev}
        \includegraphics[scale = 0.15]{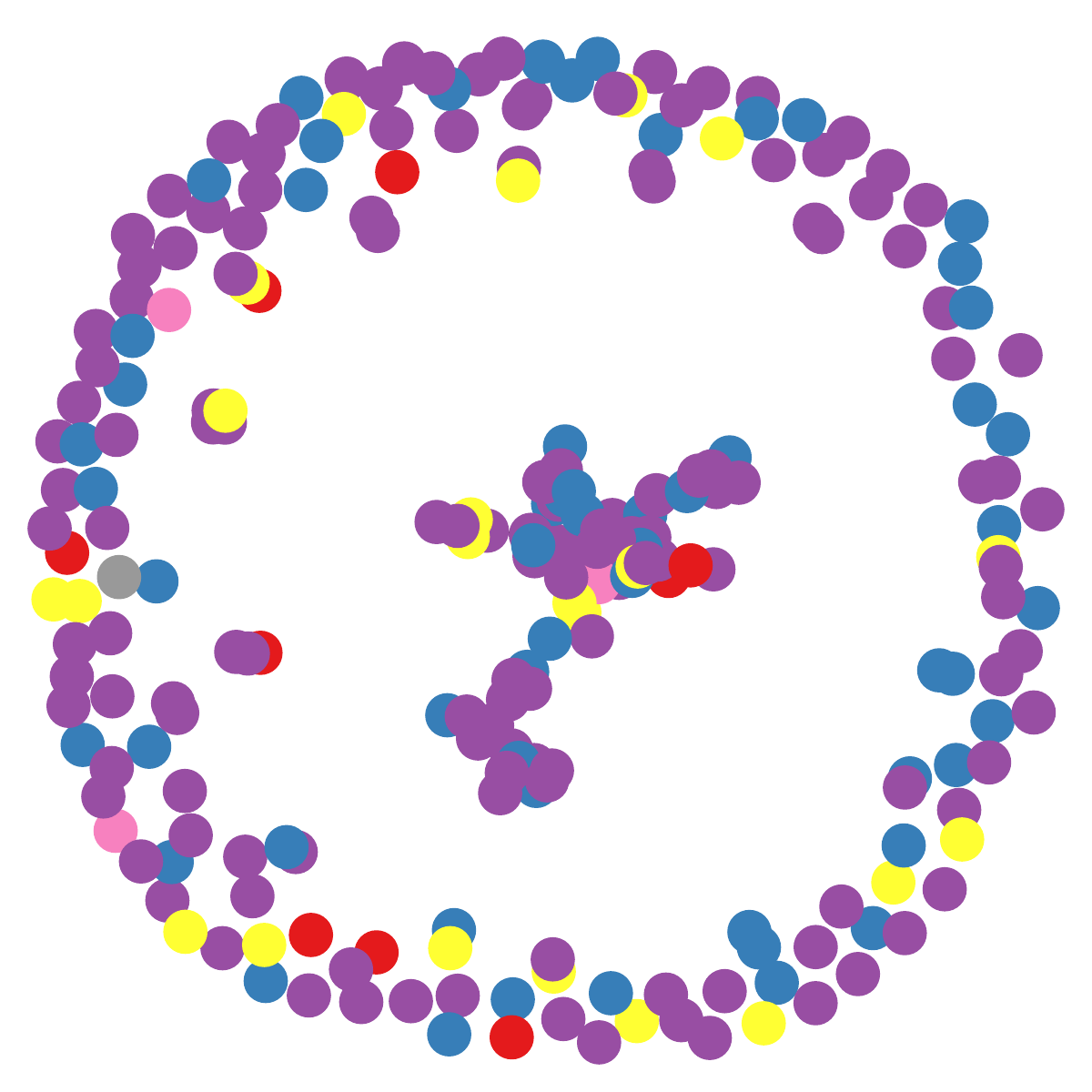}
    }
    \hspace{-4mm}
    \subfigure[Pure node sets]{
        \label{fig: Visualization pns}
        \includegraphics[scale = 0.15]{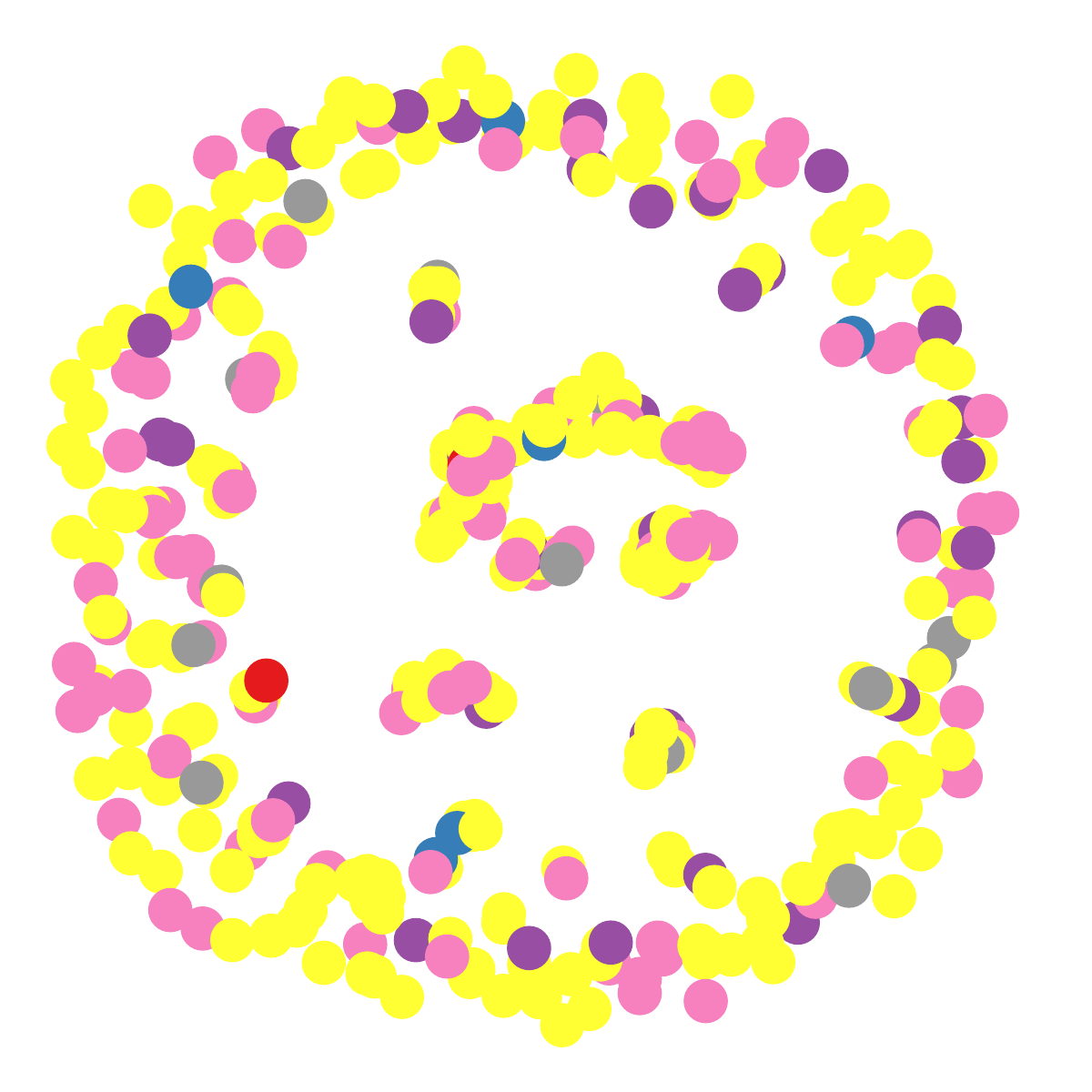}
    }
    \caption{Visualization}
    \label{fig: Visualization}
\end{figure}

\subsection{Visualization}

This section presents a visualization of the node distribution after applying pure node sampling, as shown in Figure \ref{fig: Visualization}. In this visualization, we present the original CiteSeer dataset, alongside the node distribution after applying the PNS method, which yields two distinct sets of nodes: pure nodes, filtered by PNS; and confusion nodes, identified as potentially noisy or ambiguous. Note that different colors represent different classes of nodes. To emphasize the nodes, we have omitted the depiction of edges in this visualization. As illustrated in Figure \ref{fig: Visualization ori}, the left figure displays the node distribution of the original CiteSeer dataset, which comprises six distinct classes, denoted by the colors red, purple, pink, yellow, grey, and blue; the middle Figure \ref{fig: Visualization dev} depicts the confusion node set, which is excluded from the training phase. As the figure shows, the majority of these nodes belong to the purple and blue classes, indicating that most nodes in these classes have improbable neighbor label distributions and are therefore not utilized in the node synthesis process; the right Figure \ref{fig: Visualization pns} represents the pure node set, which comprises the remaining nodes after the PNS method. As the figure shows, the majority of these nodes belong to the yellow class. The pure node set serves as the final training set, utilized for both node synthesis and model training.

\subsection{Influence of Imbalance Ratio}
\begin{figure}
    \centering
    \includegraphics[scale=0.35]{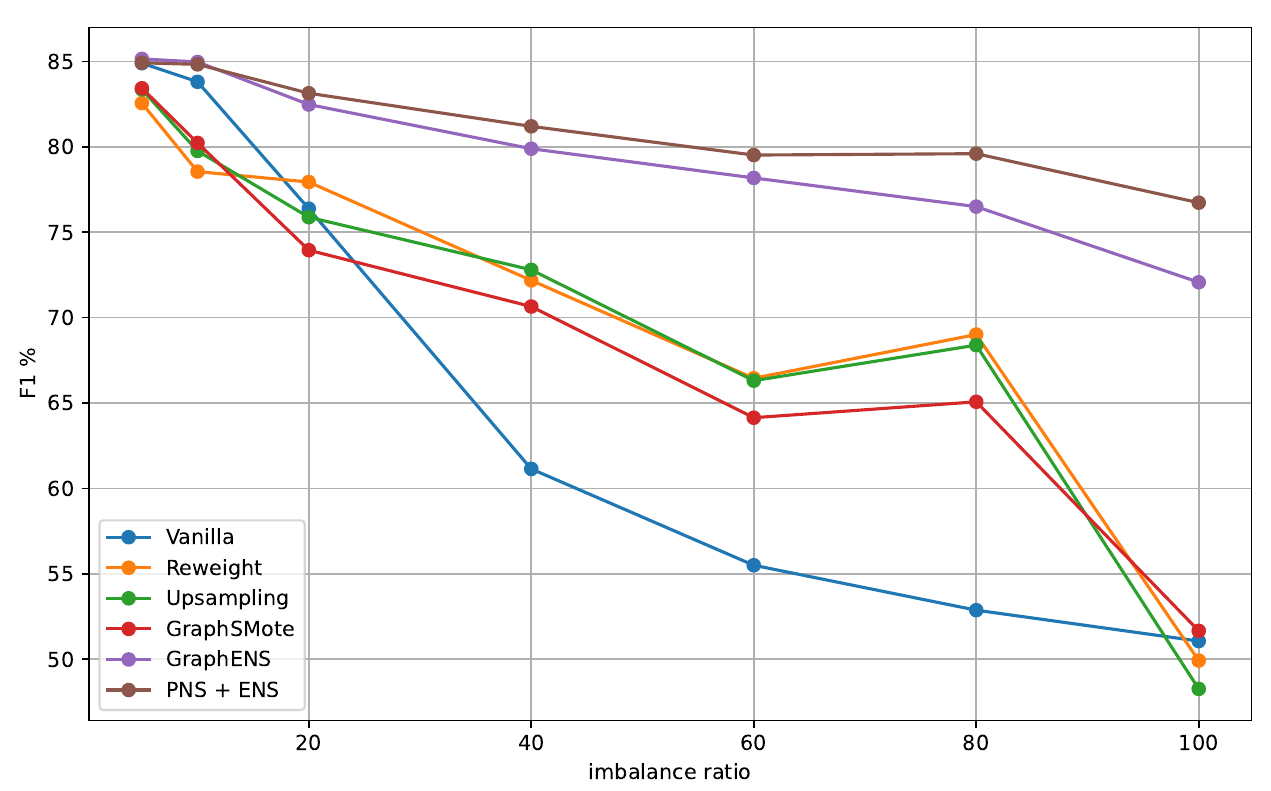}
    \caption{Changing trend of F1-score with the increase of imbalance ratio on PubMed-LT with GCN.}
    \label{fig: imbalance ratio change}
\end{figure}

Furthermore, we analyze the performance of various baselines under different imbalance ratios $\rho$, ranging from 5 to an extreme condition of 100, as illustrated in Fig. \ref{fig: imbalance ratio change} on PubMed-LT with GCN backbone. The F1 scores of all methods are consistently high when $\rho$ is relatively small. Notably, as $\rho$ increases, the performance of PNS remains remarkably stable, demonstrating its effectiveness in addressing extreme class imbalance problems on graph-structured data.

\subsection{In-depth analysis with GraphENS and TAM } \label{sec: comparison}

\begin{table*}[t]
\footnotesize
\setlength{\tabcolsep}{1.5 pt}
\renewcommand\arraystretch{1.5}
\caption{Comparison between GraphENS, TAM, and PNS.  We set the imbalance ratio in an extreme setting ($\rho$ = 80) with standard errors five times.}
\label{table: Comparision}
\begin{tabular}{c|cccc|ccc|ccc}
\hline
\multicolumn{1}{l|}{} &
  \multicolumn{1}{l}{} &
  \multicolumn{3}{c|}{\textbf{GraphENS}} &
  \multicolumn{3}{c|}{\textbf{TAM}} &
  \multicolumn{3}{c}{\textbf{PNS}} \\ \hline
 &
  Dataset &
  Acc. &
  BAcc. &
  F1 &
  Acc. &
  BAcc. &
  F1 &
  Acc. &
  BAcc. &
  F1 \\ \hline
\multirow{3}{*}{\rotatebox{90}{GCN}} &

  Photo &
  81.17 \scriptsize$\pm$ 0.38 &
  83.40 \scriptsize$\pm$ 0.42 &
  78.64 \scriptsize$\pm$ 0.90 &
  \textbf{83.61 \scriptsize$\pm$ 0.24} &
  \textbf{83.84 \scriptsize$\pm$ 0.23} &
  \textbf{81.08 \scriptsize$\pm$ 0.31} &
  82.01 \scriptsize$\pm$ 0.67 &
  83.34 \scriptsize$\pm$ 0.68 &
  80.13 \scriptsize$\pm$ 0.76 \\
 &
  Computer &
  75.36 \scriptsize$\pm$ 0.27 &
  82.72 \scriptsize$\pm$ 0.16 &
  69.72 \scriptsize$\pm$ 0.35 &
  72.97 \scriptsize$\pm$ 0.23 &
  81.23 \scriptsize$\pm$ 0.11 &
  71.11 \scriptsize$\pm$ 0.28 &
  \textbf{76.08 \scriptsize$\pm$ 0.56} &
  \textbf{82.65 \scriptsize$\pm$ 0.31} &
  \textbf{70.87 \scriptsize$\pm$ 0.26} \\
 &
  CS &
  85.05 \scriptsize$\pm$ 0.54 &
  85.45 \scriptsize$\pm$ 0.30 &
  74.68 \scriptsize$\pm$ 0.34 &
  \textbf{89.53 \scriptsize$\pm$ 0.26} &
  \textbf{88.12 \scriptsize$\pm$ 0.11} &
  \textbf{77.99 \scriptsize$\pm$ 0.18} &
  87.49 \scriptsize$\pm$ 0.52 &
  87.09 \scriptsize$\pm$ 0.25 &
  76.68 \scriptsize$\pm$ 0.12 \\ \hline
\multirow{3}{*}{\rotatebox{90}{GAT}} &

  Photo &
  83.45 \scriptsize$\pm$ 0.20 &
  85.93 \scriptsize$\pm$ 0.38 &
  82.08 \scriptsize$\pm$ 0.29 &
  64.69 \scriptsize$\pm$ 0.72 &
  63.08 \scriptsize$\pm$ 1.14 &
  62.15 \scriptsize$\pm$ 0.75 &
  \textbf{83.78 \scriptsize$\pm$ 0.35} &
  \textbf{85.19 \scriptsize$\pm$ 0.24} &
  \textbf{82.47 \scriptsize$\pm$ 0.27} \\
 &
  Computer &
 76.20 \scriptsize$\pm$ 0.50 &
  \textbf{83.63\scriptsize$\pm$ 0.34 }&
  71.26 \scriptsize$\pm$ 0.65 &
  67.99 \scriptsize$\pm$ 1.12 &
  75.02 \scriptsize$\pm$ 0.29 &
  63.56 \scriptsize$\pm$ 1.55 &
  \textbf{76.75 \scriptsize$\pm$ 0.25} &
  83.38 \scriptsize$\pm$ 0.12 &
  \textbf{72.19 \scriptsize$\pm$ 0.31} \\
 &
  CS &
  87.54 \scriptsize$\pm$ 0.31 &
  87.81 \scriptsize$\pm$ 0.28 &
  76.17 \scriptsize$\pm$ 1.41 &
  87.80 \scriptsize$\pm$ 0.37 &
  85.13 \scriptsize$\pm$ 0.32 &
  73.71 \scriptsize$\pm$ 1.28 &
  \textbf{89.31 \scriptsize$\pm$ 0.46} &
  \textbf{88.74 \scriptsize$\pm$ 0.19} &
  \textbf{78.97 \scriptsize$\pm$ 0.35} \\ \hline
\multirow{3}{*}{\rotatebox{90}{SAGE}} &

  Photo &
  81.87 \scriptsize$\pm$ 0.38 &
  83.40 \scriptsize$\pm$ 0.42 &
  78.64 \scriptsize$\pm$ 0.90 &
  80.15 \scriptsize$\pm$ 0.55 &
  83.35 \scriptsize$\pm$ 0.45 &
  76.25 \scriptsize$\pm$ 0.53 &
  \textbf{82.06 \scriptsize$\pm$ 1.08} &
  \textbf{83.36 \scriptsize$\pm$ 0.71} &
  \textbf{79.80 \scriptsize$\pm$ 1.02} \\
 &
  Computer &
  72.72 \scriptsize$\pm$ 0.57 &
  80.19 \scriptsize$\pm$ 0.28 &
  66.26 \scriptsize$\pm$ 0.57 &
  65.58 \scriptsize$\pm$ 0.74 &
  72.82 \scriptsize$\pm$ 0.45 &
  61.79 \scriptsize$\pm$ 1.14 &
  71.80 \scriptsize$\pm$ 0.56&
  76.94 \scriptsize$\pm$ 0.69
  &
  64.17 \scriptsize$\pm$ 1.01 \\
 &
  CS &
  85.61 \scriptsize$\pm$ 0.55 &
  86.53 \scriptsize$\pm$ 0.24 &
  75.05 \scriptsize$\pm$ 0.74 &
  \textbf{87.53 \scriptsize$\pm$ 0.46} &
  86.92 \scriptsize$\pm$ 0.15 &
  \textbf{76.71 \scriptsize$\pm$ 0.26} &
  87.52 \scriptsize$\pm$ 0.37 &
  \textbf{87.25 \scriptsize$\pm$ 0.14} &
  75.96 \scriptsize$\pm$ 1.07 \\ \hline
\end{tabular}

\end{table*}

This section presents a comparative analysis of PNS, GraphENS, and TAM, highlighting their differences and distinct characteristics. Furthermore, Table \ref{table: Comparision} presents a comparison of the performance of GraphENS, TAM, and PNS across three diverse datasets. The table uses GraphENS as a baseline, allowing for a direct comparison of the relative performance of TAM and PNS. Noticed that we set $\rho$ = 80 in Photo, Computer, and CS.

\textbf{Comparisons with GraphENS}. GraphENS considers the impact of message passing between nodes and hypothesizes that overfitting to the neighbor sets of minority classes is a major challenge for class-imbalanced node classification. It synthesizes the entire ego network for the minority class by combining two diverse ego networks, selected based on their similarity to the minority node and its one-hop neighbors. One limitation of GraphENS is that it overlooks the anomalous connectivity issue in some nodes and their corresponding ego networks, which can mislead the GNN model during message passing.  PNS addresses this issue by proactively filtering out problematic nodes, thereby synthesizing nodes with reduced noise. As shown in Table \ref{table: Comparision}, PNS consistently outperforms GraphENS across all datasets and GNN backbones. Notably, PNS achieves substantial performance gains in all datasets except Photo, where the improvement is less pronounced. Using GCN as the backbone, PNS achieves notable improvements in accuracy, balanced accuracy, and F1 score across various datasets. Specifically, it yields accuracy gains of 2.44\% on CS. Additionally, it obtains balanced accuracy improvements of 1.64\%, and F1 score gains 2.00\% on these datasets. In summary, GraphENS fails to account for the anomalous connectivity issue in certain nodes and their corresponding ego networks. Furthermore, it may also be susceptible to the RACP issue, which can lead to the selection of nodes with AC problems for synthesis.

\textbf{Comparisons with TAM}. TAM has identified the issue of anomalous connectivity. It hypothesizes that the label distribution surrounding each node has a significant impact on the increase in false positive cases, and this hypothesis is empirically validated through experiments. TAM detects nodes with topologically anomalous connections and adaptively refines the network structure by adjusting their margins. It presents two innovative margin adaptation strategies: the anomalous connectivity-aware margin (ACM), which dynamically adjusts the margin for each class, and the anomalous distribution-aware margin (ADM), which adaptively refines the target class margin based on its relative proximity to the target class versus the self-class. TAM integrates both ACM and ADM to refine the model's learning objective. In contrast, PNS offers a more direct and efficient solution to anomalous connectivity issues, simultaneously mitigating RACP problems and improving performance. As shown in Table \ref{table: Comparision}, TAM surpasses PNS in performance on Photo and CS when using GCN as the GNN backbone. However, PNS achieves superior accuracy and balanced accuracy on Computer, with improvements of 3.11\% in Acc. and 1.42\% in bAcc. When using GAT as the GNN backbone, PNS outperforms TAM on Photo, Computer, and CS. TAM suffers from significant performance degradation on Photo and Computer, with accuracy drops of 18.98\% and 14.04\%, balanced accuracy declines of 20.08\% and 8.68\%, and F1 score decreases of 20.02\% and 16.05\%, respectively. When using SAGE, PNS achieves marginally better performance on Photo and Computer. PNS performs slightly worse than TAM on CS, with accuracy declines of 0.10\%, and F1 score drops of 0.70\%, respectively. In summary, PNS outperforms TAM when using GAT and SAGE, but is outperformed by TAM when using GCN as the backbone.

\textbf{Why does PNS perform better?} Most of the methods with node synthesis ignore the RACP during the node sample stage. With the sampling node having an anomalous connectivity issue, the synthesized new node would tend to potentially have an anomalous connectivity issue and an ambiguous feature. Then in the message-passing mechanism $h_v^{(l)} = \texttt{UPDATE}\left(h_v^{(l-1)},\ \texttt{AGG}\left(\left\{ h_u^{(l-1)} \mid u \in \mathcal{N}(v) \right\}\right)\right)$, both $h_v^{(l-1)}$  and $\texttt{AGG}\left(\left\{ h_u^{(l-1)} \mid u \in \mathcal{N}(v) \right\}\right)$ can’t represent the the class information, especially after update the representation with them. PNS filters out some of the confusion nodes that with high levels of anomalous connectivity issues before embedding, which optimizes  $\texttt{AGG}\left(\left\{ h_u^{(l-1)} \mid u \in \mathcal{N}(v) \right\}\right)$ in message-passing, resulting in a better node representation. Due to this mechanism, PNS can even mitigate performance degradation caused by the abnormal distribution of nodes.

\subsection{Complexity Analysis}

In this subsection, we analyze the time and GPU memory consumption in the forward propagation of our proposed method. We compare its performance with that of other models, including Vanilla, which is used as a baseline,  and GraphENS, to provide a comprehensive evaluation of its efficiency and scalability. We analyze the time and GPU memory complexity of three GNN backbones, namely GCN, GAT, and SAGE, across six different benchmarks: Cora, CiteSeer, PubMed, Photo, Computer, and CS.

As shown in Table \ref{table: complexity}, the computational complexity of the GNN backbones is as follows: Methods based on GCN exhibit the highest time consumption on the computer dataset and the highest GPU memory usage on the CS dataset. Notably, GraphENS and PNS require significantly more time than Vanilla, with increases of 1 m 22 s and 1 m 19 s, respectively. Additionally, GraphENS and PNS consume substantially more memory than Vanilla, with excess memory usage of 1801.17 MB and 1798.59 MB, respectively. Similarly, in GAT, the time and memory consumption patterns mirror those of GCN, with models requiring the most time on the Computer dataset and the most memory on the CS dataset. Specifically, on the Computer dataset, GraphENS and PNS incur additional time costs of 49 seconds and 51 seconds, respectively, compared to Vanilla. On the CS dataset, GraphENS and PNS incur additional time costs of 1 minute 27 seconds and 1 minute 25 seconds, respectively, compared to Vanilla, and also consume 1801.16 MB and 1799.16 MB more memory than Vanilla. In contrast to the above GNN backbones, SAGE exhibits distinct conditions, with GraphENS and PNS requiring the most time and memory on the CS dataset. Specifically, they incur additional time costs of 4 minutes 12 seconds and 4 minutes 22 seconds, respectively, compared to Vanilla. Furthermore, ENS and PNS also consume 1801.71 MB and 1799.72 MB more memory than Vanilla on this dataset. Finally, we analyze the time complexity of our proposed method. Let $o$ denote the complexity of the base model, $v$ represents the total number of nodes, and $n$ represents the average number of neighbors per node. The time complexity of constructing the NLD is $O(v)$, as it involves a single pass through all nodes. The time complexity of filtering out confusion nodes is $O(nv)$, since we must traverse the neighbors of all nodes. Therefore, the overall time complexity of our method is $O(v(n+1)+o)$, which accounts for both the node-based and neighbor-based operations.

In summary, our experiments reveal that the models incur the highest time and memory costs on the PubMed and CS datasets.  GraphENS and PNS require significantly more time and memory compared with Vanilla. Furthermore, PNS is more efficient than GraphENS, consuming less time and memory.

\subsection{Limitations} \label{sec: limitation}

While PNS achieves superior performance compared to various baselines in our experiments, we also acknowledge and reflect on its limitations. 1) When data samples are extremely limited, PNS may exacerbate the scarcity issue, resulting in even fewer available samples for subsequent node synthesis tasks. This can lead to severe homogenization of the synthesized samples, potentially causing underrepresentation issues in downstream models. 2) The confusion rate $r$ must be appropriately set. Otherwise, all nodes in a dataset would be judged as confusion nodes, then no nodes would remain for synthesis. Conversely, if the dataset consists solely of pure nodes, with no confusion nodes present, the effectiveness of PNS will be lost. However, typical datasets often contain a substantial amount of noise, which manifests as confusion nodes. 3) PNS determines whether a node is pure based on a single-dimensional purity measure,  effectively excluding nodes with anomalous connectivity (AC) and decoupling node selection from random seeds. However, relying solely on purity may be insufficient to fully capture the heterogeneity of a node’s neighborhood.

\begin{table*}[h]
\setlength{\tabcolsep}{4pt}
\renewcommand\arraystretch{1.2}
\caption{Complexity of different models.}
\label{table: complexity}
\begin{tabular}{l|l|ll|ll|ll}

\hline
                      &          & \multicolumn{2}{c|}{\textbf{Vanilla}} & \multicolumn{2}{c|}{\textbf{GraphENS}} & \multicolumn{2}{c}{\textbf{PNS}} \\ \hline
\multicolumn{1}{c|}{\textbf{Backbone}} &
  \multicolumn{1}{c|}{\textbf{Dataset}} &
\multicolumn{1}{c}{ \makecell{\textbf{Time} \\ \textbf{Consumption}}} &
  \multicolumn{1}{c|}{\makecell{\textbf{Memory} \\ \textbf{Consumption}}} &
\multicolumn{1}{c}{ \makecell{\textbf{Time} \\ \textbf{Consumption}}} &
  \multicolumn{1}{c|}{\makecell{\textbf{Memory} \\ \textbf{Consumption}}} &
\multicolumn{1}{c}{ \makecell{\textbf{Time} \\ \textbf{Consumption}}} &
  \multicolumn{1}{c}{\makecell{\textbf{Memory} \\ \textbf{Consumption}}}  \\ \hline
\multirow{6}{*}{GCN}  & Cora     & 00h 00m 35s     & 6.67 MB    & 00h 01m 00s    & 60.14 MB     & 00h 01m 01s   & 59.69 MB     \\
                      & CiteSeer & 00h 00m 36s     & 15.54MB    & 00h 01m 11s    & 127.84MB     & 00h 01m 09s   & 124.77 MB    \\
                      & PubMed   & 00h 01m 02s     & 3.28 MB    & 00h 05m 38s    & 1511.28MB    & 00h 05m 13s   & 1547.10 MB   \\
                      & Photo    & 00h 01m 56s     & 3.99MB     & 00h 02m 24s    & 251.09MB     & 00h 02m 19s   & 251.22 MB    \\
                      & Computer & 00h 03m 33s     & 4.11MB     & 00h 04m 17s    & 769.77 MB    & 00h 04m 20s   & 769.99 MB    \\
                      & CS       & 00h 02m 43 s    & 27.74 MB   & 00h 04m 05s    & 1828.91 MB   & 00h 04m 02s   & 1826.33 MB   \\ \hline
\multirow{6}{*}{GAT}  & Cora     & 00h 00m 32s     & 6.68 MB    & 00h 00m 55s    & 60.15 MB     & 00h 00m 59s   & 59.71 MB     \\
                      & CiteSeer & 00h 00m 33s     & 15.55 MB   & 00h 01m 05s    & 127.85 MB    & 00h 01m 07s   & 124.79 MB    \\
                      & PubMed   & 00h 01m 15s     & 3.29 MB    & 00h 06m 04s    & 1511.30 MB   & 00h 05m 42s   & 1547.12 MB   \\
                      & Photo    & 00h 02m 16s     & 4.00 MB    & 00h 02m 41s    & 251.10 MB    & 00h 02m 41s   & 251.24 MB    \\
                      & Computer & 00h 04m 26s     & 4.13 MB    & 00h 05m 15s    & 769.79 MB    & 00h 05m 17s   & 770.00 MB    \\
                      & CS       & 00h 03m 01s     & 27.76 MB   & 00h 04m 28s    & 1828.92 MB   & 00h 04m 26s   & 1826.92 MB   \\ \hline
\multirow{6}{*}{SAGE} & Cora     & 00h 00m 27s     & 13.26 MB   & 00h 00m 53s    & 66.14 MB     & 00h 00m 53s   & 65.52 MB     \\
                      & CiteSeer & 00h 00m 38s     & 31.00 MB   & 00h 01m 20s    & 143.30 MB    & 00h 01m 18s   & 141.21 MB    \\
                      & PubMed   & 00h 00m 54s     & 6.23 MB    & 00h 05m 23s    & 1554.23 MB   & 00h 05m 11s   & 1550.06 MB   \\
                      & Photo    & 00h 01m 46s     & 7.90 MB    & 00h 02m 26s    & 255.00 MB    & 00h 02m 23s   & 255.13 MB    \\
                      & Computer & 00h 03m 31s     & 8.11 MB    & 00h 04m 49s    & 773.77 MB    & 00h 04m 55s   & 773.98 MB    \\
                      & CS       & 00h 10m 19s     & 55.33 MB   & 00h 14m 31s    & 1857.04 MB   & 00h 14m 41s   & 1855.05 MB   \\ \hline
\end{tabular}

\end{table*}

\section{In-depth analysis of RACP}
\label{sec: In-depth analysis of RACP}
In this section, we make a series of experiments to analyze RACP in-depth. First of all, we reproduce RACP and compare the performance between it and non-RACP. Then we analyze its cause based on the theory in Section \ref{sec: Theoretical Analysis of RACP}, Finally, we explore RACP in different models.
\subsection{Experiment of Reproduce RACP}

\textbf{Arrangement.} We hypothesize that an unlucky random seed tends to select nodes with anomalous connectivity. We verify our hypothesis by selecting nodes only with anomalous connectivity (we call it \textit{confusion node} below) compared with selecting nodes without any anomalous connectivity (we call it \textit{pure node} below). Initially, we calculate the neighbor label distribution (NLD) of each node. Next, we want to simulate an extreme situation. We only select the confusion node or the pure node to mix up. Our experiment was performed with GraphENS \citep{park2021graphens} to verify the randomness anomalous connectivity problem because AC without randomness has already been proven in TAM \citep{song2022tam}.

\textbf{Settings.} We conduct experiments on three benchmarks: Cora, Citseer, and PubMed \citep{sen2008collective}. Expect only to use the \textit{confusion node} and \textit{pure node} to mix up, all settings are followed by GraphENS' semi-supervised experiment \citep{park2021graphens}. It is worth noting that we adopt a full data split on Cora, PubMed, and CiteSeer following the way with GraphENS. In the semi-supervised experiment of reproducing RACP, we conduct the original experiment in GraphENS. We set the imbalance ratio $\rho$ = 10, confusion rate $r$ = 0.8, and only select the confusion node to synthesize.

\textbf{Results.} Our experiment demonstrates that the selected node is a significant factor influenced by randomness, like in Fig. \ref{fig: confusion compare}. The unlucky seed significantly impaired the performance of GrapnENS, and our experiment in extreme situations was even more disappointing. Notice that after our debugging, we found GraphENS with a bad seed selecting 40\% confusion minor node to synthesize. We observe that the model's performance significantly deteriorates when only the confusion nodes are used for synthesis. In contrast, using only the \textit{pure node} for synthesis maintains strong performance. In the following experiment, we can see that pure node sampling not only addresses the randomness anomalous connectivity problem but also improves overall model performance. 

\textbf{How to reproduce RACP?} We can observe that the RACP problem is in the semi-supervised node classification task of GraphENS. As illustrated in Figure \ref{fig: rac problem pic}, we observe two distinct scenarios:  (1) when the seed is set to 100 (denoted as ``normal''), GraphENS exclusively selects pure nodes for synthesis; and (2) when the seed is set to 102 (denoted as ``bad seed''), GraphENS inadvertently selects a portion of confusion nodes for synthesis, resulting in a decline in performance. To simplify the experimental setup, in the ``worst seed'' scenario, we only consider the confusion nodes for the synthesis process. Our experimental results reveal that the choice of random seed can significantly impact the selection of nodes for synthesis during the node synthesis stage, ultimately influencing the model's performance. For a more detailed reproduction of our experiments, please refer to https://github.com/flzeng1/PNS.

\subsection{Analysis of RACP}

This section analyzes the impact of the randomness anomalous connectivity problem (RACP) on the performance of GNN models in imbalanced node classification tasks. In general, synthesizing nodes for the minority class is necessary to balance the training set and mitigate class imbalance. A crucial step in this process is the judicious selection of nodes for synthesis. Numerous methods have investigated this process, such as GraphENS, which leverages not only individual nodes but also their ego networks to synthesize minority nodes. However, a critical oversight of these methods is that the selection of nodes for synthesis is often random and unguided. For instance, GraphENS samples target nodes from a multinomial distribution, which can inadvertently include nodes with the anomalous connectivity (AC) issue. Utilizing nodes with the AC problems to synthesize minority nodes can lead to an increase in false positive cases, as these nodes often exhibit topologically improbable connections. We consider a scenario in which a random seed is used to select a mix of nodes, roughly evenly split between normal nodes and nodes with AC problems, for synthesizing minority nodes, resulting in average model performance. In summary, if a random seed selects only nodes with AC problems for minority node synthesis, it degrades model performance. Conversely, if the seed selects only normal nodes, it leads to a performance above the average.

\subsection{RACP in Different Models}

\begin{table*}[]
\caption{RACP in different models. Ori denotes the original performance; Pns denotes the performance that only uses the pure node; Dev denotes the performance that only uses the confusion node.}
\label{table: RACP in Different Model}
\renewcommand\arraystretch{1.2}
\begin{tabular}{l|l|c|c|c|c|c|c|c}
\hline
 &
   &
  Vanilla &
  Reweight &
  Upsampling &
  GraphSMOTE &
  GraphENS &
  PC Softmax &
  GraphSHA \\ \hline
\multirow{3}{*}{Ori} &
  Acc. &
  35.35 \scriptsize$\pm$ 0.15 &
  38.77 \scriptsize$\pm$ 0.06 &
  38.78 \scriptsize$\pm$ 1.57 &
  39.75 \scriptsize$\pm$ 1.65 &
  83.42 \scriptsize$\pm$ 1.03 &
  38.71 \scriptsize$\pm$ 0.61 &
  86.63 \scriptsize$\pm$ 0.24 \\
 &
  BAcc. &
  48.37 \scriptsize$\pm$ 0.16 &
  50.23 \scriptsize$\pm$ 0.91 &
  49.79 \scriptsize$\pm$ 0.63 &
  50.59 \scriptsize$\pm$ 0.30 &
  82.42 \scriptsize$\pm$ 0.11 &
  49.91 \scriptsize$\pm$ 0.51 &
  87.04 \scriptsize$\pm$ 0.13 \\
 &
  F1 &
  22.02 \scriptsize$\pm$ 1.55 &
  24.96 \scriptsize$\pm$ 0.04 &
  24.68 \scriptsize$\pm$ 0.07 &
  24.06 \scriptsize$\pm$ 0.62 &
  73.74 \scriptsize$\pm$ 0.77 &
  34.53 \scriptsize$\pm$ 0.24 &
  75.86  \scriptsize$\pm$0.35 \\ \hline
\multirow{3}{*}{Pns} &
  Acc. &
  63.28 \scriptsize$\pm$ 0.71 &
  70.95 \scriptsize$\pm$ 2.09 &
  58.32 \scriptsize$\pm$ 10.82 &
  70.25 \scriptsize$\pm$ 0.62 &
  87.49 \scriptsize$\pm$ 0.52 &
  89.13 \scriptsize$\pm$ 0.09 &
  87.06 \scriptsize$\pm$ 0.19 \\
 &
  BAcc. &
  64.49 \scriptsize$\pm$ 0.28 &
  73.31 \scriptsize$\pm$ 1.03 &
  63.74 \scriptsize$\pm$ 7.32 &
  73.09 \scriptsize$\pm$ 0.56 &
  87.09 \scriptsize$\pm$ 0.25 &
  87.04 \scriptsize$\pm$ 0.78 &
  87.68 \scriptsize$\pm$ 0.15 \\
 &
  F1 &
  46.74 \scriptsize$\pm$ 0.55 &
  56.91 \scriptsize$\pm$ 0.15 &
  44.37 \scriptsize$\pm$ 8.89 &
  58.85 \scriptsize$\pm$ 0.10 &
  76.34 \scriptsize$\pm$ 0.17 &
  81.23 \scriptsize$\pm$ 0.12 &
  76.46 \scriptsize$\pm$ 0.49 \\ \hline
\multirow{3}{*}{Dev} &
  Acc. &
  32.31 \scriptsize$\pm$ 0.33 &
  31.62 \scriptsize$\pm$ 2.30 &
  30.26 \scriptsize$\pm$ 0.18 &
  - &
  70.44 \scriptsize$\pm$ 1.44 &
  5.40 \scriptsize$\pm$ 0.00 &
  87.30 \scriptsize$\pm$ 0.21 \\
 &
  BAcc. &
  46.02 \scriptsize$\pm$ 0.31 &
  43.23 \scriptsize$\pm$ 1.17 &
  43.22 \scriptsize$\pm$ 0.65 &
  - &
  75.79 \scriptsize$\pm$ 0.74 &
  7.14 \scriptsize$\pm$ 0.00 &
  87.51 \scriptsize$\pm$ 0.33 \\
 &
  F1 &
  20.47 \scriptsize$\pm$ 0.51 &
  18.40 \scriptsize$\pm$ 0.95 &
  17.77 \scriptsize$\pm$ 0.29 &
  - &
  59.13 \scriptsize$\pm$ 1.67 &
  7.30 \scriptsize$\pm$ 0.00 &
  76.13 \scriptsize$\pm$ 0.42 \\ \hline
\end{tabular}

\end{table*}

We also explore whether other models meet the same RACP problem. We use Vanilla, Reweight, Upsampling, graphSMOTE, graphENS, PC Softmax, and GraphSHA as the baseline in the CS dataset. Notice that Ori denotes the original performance; Pns denotes the performance that only uses pure nodes; Dev denotes the performance that only uses confusion nodes. We repeat all experiments two times with the same parameter settings. Why don't we select the bad seeds and instead focus on confusing nodes? It is because it would be a waste of time. Instead, we only consider confusing nodes to conduct the experiment, which serves as a potential lower bound, representing a worst-case scenario.

As shown in Table \ref{table: RACP in Different Model}, we can see that except GraphSHA, others exhibit performance degradation of varying degrees with confusion nodes. Additionally, except GraphSHA, others emerge to greatly improve performance with pure nodes. As Vanilla, Reweight, Upsampling, PC softmax have gained tremendous performance improvement, with increases of performance 27.93\%, 32.18\%, 19.54\%, 30.50\%, 50.42\% in Accuracy; 16.12\%, 23.08\&, 13.95\%, 22.50\%, 37.13\% in BAcc. As for graphENS, also gained the improved performance, with increments of 2.84\% in Acc, 2.59\%, and 3.17\%. In contrast, GraphSHA has hardly changed. In terms of confusion nodes,  except GraphSHA, others emerge with great performance degradation. Such as Vanilla, Reweight, Upsampling, PC softmax, with the performance 3.04$\% \downarrow$, 7.15$\% \downarrow$, 8.52$\% \downarrow$, 33.31$\% \downarrow$ in Acc; 2.35$\% \downarrow$, 7.00$\% \downarrow$, 6.57$\% \downarrow$, 42.77$\% \downarrow$ in Bacc, etc. As for graphENS, also meet a sharp performance degradation, with the performance of 12.98$\% \downarrow$ in Acc., 6.63$\% \downarrow$ in BAcc., and 14.61$\% \downarrow$ in F1. In contrast, GraphSHA has hardly changed.

In summary, GraphSHA has hardly changed its performance in the experiment, which means GraphSHA may not have any issues with RACP. The performance of other methods will be greatly affected by the selection of node synthesis, which means these methods most likely encounter RACP issues.

\subsection{Difference between RACP and AC}

Anomalous connectivity (AC) refers to the phenomenon in which minor nodes that deviate from the dominant connectivity pattern generate an excessive number of false positives during the quantity-based compensation process. AC arises when incorporating such minor nodes leads to the synthesis of additional nodes that also deviate from the pattern, ultimately resulting in performance degradation. TAM employs ACM and ADM to dynamically adjust the loss weights of nodes exhibiting AC during the learning process. The randomness anomalous connectivity problem (RACP) arises during node synthesis, where the model randomly selects nodes with AC and synthesizes minor nodes with varying random seeds. RACP is a probabilistic issue, meaning it may not occur consistently, but its likelihood of occurrence is non-zero and dependent on the random seed used, introducing unpredictability into the node synthesis process. PNS introduces the concept of a confusion rate, which enables the filtering of nodes affected by the AC problem, referred to as confusion nodes. Existing methods often overlook the crucial step of node selection during the synthesis process. In contrast, TAM addresses this issue through a distinct approach, modifying the loss weight for these nodes to mitigate the problem. Meanwhile, PNS directly filters out a subset of these nodes.

\section{Conclusion} \label{sec: Conclusion}

In this paper, we explore the graph imbalance node classification task and uncover the randomness-induced anomalous connectivity problem inherent in GraphENS. To address this issue, we conduct a set of experiments to identify the factors affected by the random seed. We subsequently propose PNS, tailored to mitigate the randomness-induced anomalous connectivity problem. By leveraging PNS, we effectively eliminate the randomness problem and narrow the sampling boundary for node synthesis, thereby reducing confusion in GNN backbones. PNS mitigates performance degradation caused by the abnormal distribution of node neighbors. Experimental results demonstrate that PNS consistently outperforms various baselines across different datasets, showcasing its robust performance on classical GNN backbones. In future work, we plan to further investigate the RACP problem in the node synthesis process. Our future research directions include developing an automated method for selecting the optimal confusion rate to enhance PNS effectiveness.

\section*{Acknowledgment}

This research was supported in part by National Natural Science Foundation of China (No. 62272196), the Guangzhou Basic and Applied Basic Research Foundation (No. 2024A04J9971).

\section*{Data Availability}
Data and code are available at https://github.com/flzeng1\\/PNS.

\section*{CRediT Authorship Contribution Statement}
Fanlong Zeng: Methodology, writing original draft.
Wensheng Gan: Review and editing, supervision.
Jiayang Wu: Visualization.
Philip S. Yu: Review and editing.

\bibliographystyle{apalike}
\bibliography{pns.bib}

\end{document}